\documentclass{article}

\usepackage{amsmath}
\usepackage{mathtools}
\usepackage{amsfonts}

\usepackage{microtype}
\usepackage{graphicx}
\usepackage{subfigure}
\usepackage{booktabs} % for professional tables

\usepackage{hyperref}

\usepackage[accepted]{icml2019}

\usepackage[symbol]{footmisc}

\icmltitlerunning{Discretizing Continuous Action Space for On-Policy Optimization}

\begin{document}

\twocolumn[
\icmltitle{Discretizing Continuous Action Space for On-Policy Optimization}




\begin{icmlauthorlist}
\icmlauthor{Yunhao Tang}{columbia}
\icmlauthor{Shipra Agrawal}{columbia}
\end{icmlauthorlist}

\icmlaffiliation{columbia}{Columbia University, New York, NY, USA}

\icmlcorrespondingauthor{Yunhao Tang}{yt2541@columbia.edu}

\icmlkeywords{Machine Learning, ICML}

\vskip 0.3in
]



\printAffiliationsAndNotice{}  % leave blank if no need to mention equal contribution

\begin{abstract}
In this work, we show that discretizing action space for continuous control is a simple yet powerful technique for on-policy optimization. The explosion in the number of discrete actions can be efficiently addressed by a policy with factorized distribution across action dimensions. We show that the discrete policy achieves significant performance gains with state-of-the-art on-policy optimization algorithms (PPO, TRPO, ACKTR) especially on high-dimensional tasks with complex dynamics. Additionally, we show that an ordinal parameterization of the discrete distribution can introduce the inductive bias that encodes the natural ordering between discrete actions. This ordinal architecture further significantly improves the performance of PPO/TRPO. An open source implementation of this paper can be found at \url{https://github.com/robintyh1/onpolicybaselines}.
\end{abstract}

\section{Background}
In reinforcement learning (RL), the action space of conventional control tasks are usually dichotomized into either discrete or continuous \citep{brockman2016}. While discrete action space is conducive to theoretical analysis, in the context of deep reinforcement learning, their application is limited to video game playing or board game \citep{mnih2013,silver2016}. On the other hand, in simulated or real life robotics control \citep{levine2016,schulman2015}, the action space is by design continuous. Continuous control typically requires more subtle treatments, since a continuous range of control contains an infinite number of feasible actions and one must resort to parametric functions for a compact representation of distributions over actions.

Can we retain the simplicity of discrete actions when solving continuous control tasks? A straightforward solution is to discretize the continuous action space, i.e. we discretize the continuous range of action into a finite set of atomic actions and reduce the original task into a new task with a discrete action space. A common argument against this approach is that for an action space with $M$ dimensions, discretizing $K$ atomic actions per dimension leads to $M^K$ combinations of joint atomic actions, which quickly becomes intractable when $M$ increases. However, a simple fix is to represent the joint distribution over discrete actions as factorized across dimensions, so that the joint policy is still tractable. As prior works have applied such discretization method in practice \citep{openai2018learning,jaskowski2018reinforcement}, we aim to carry out a systemic study of such straightforward discretization method in simulated environments, and show how they improve upon on-policy optimization baselines.

%The contributions of our paper are as follows: (1) We carry out a systemic evaluation of the straightforward discretization strategy on a wide range of continuous control tasks in OpenAI gym \citep{brockman2016}, rllab \citep{duanxi2016} and Roboschool \citep{schulman2017proximal}, where we show that this method significantly improves upon on-policy optimization algorithms such as TRPO, ACKTR and PPO \citep{schulman2015trust,wu2017scalable,schulman2017proximal} over baselines. Since $K$ is a critical hyper-parameter for this method, we also analyze the trade-off of choosing $K$. (2) We show that the \emph{stick-breaking} parameterization, originally proposed in statistics literature for ordinal classification \citep{khan2012stick}, further significantly improves the performance of PPO/TRPO.

The paper proceeds as follows. In Section 2, we introduce backgrounds on on-policy optimization baselines (e.g. TRPO and PPO) and related work. In Section 3, we introduce the straightforward method of discretizing action space for continuous control, and analyze the properties of the resulting policies as the number atomic actions $K$ changes. In Section 4, we introduce \emph{stick-breaking} parameterization \citep{khan2012stick}, an architecture that parameterizes the discrete distributions while encoding the natural ordering between discrete actions. In Section 5, through extensive experiments we show how the discrete/ordinal policy improves upon current on-policy optimization baselines and related prior works, especially on high-dimensional tasks with complex dynamics.

\section{Background}
\subsection{Markov Decision Process}
In the standard formulation of Markov Decision Process (MDP), an agent starts with an initial state $s_0 \in \mathcal{S}$ at time $t=0$. At time $t\geq 0$, the agent is in $s_t \in \mathcal{S}$, takes an action $a_t \in \mathcal{A}$, receives a reward $r_t \in \mathbb{R}$ and transitions to a next state $s_{t+1} \sim p(\cdot|s_t,a_t)$. A policy is a mapping from state to distributions over actions $\pi: \mathcal{S} \mapsto P(\mathcal{A})$. The expected cumulative reward under a policy $\pi$ is $J(\pi) = \mathbb{E}_\pi \big[\sum_{t=0}^\infty r_t\gamma^t\big]$ where $\gamma \in [0,1)$ is a discount factor. The objective is to search for the optimal policy that achieves maximum reward $\pi^\ast = \arg\max_\pi J(\pi)$. For convenience, under policy $\pi$ we define action value function $Q^\pi(s,a) = \mathbb{E}_{\pi}\big[J(\pi)|s_0=s,a_0=a\big]$ and value function $V^\pi(s) = \mathbb{E}_\pi\big[J(\pi)|s_0=s,a_0\sim \pi(\cdot|s_0)\big]$. Also define the advantage function $A^\pi(s,a) = Q^\pi(s,a) - V^\pi(s)$.

\subsection{On-Policy Optimization}
In policy optimization, one restricts the policy search within a class of parameterized policy $\pi_\theta,\theta \in \Theta$ where $\theta$ is the parameter and $\Theta$ is the parameter space. A straightforward way to update policy is to do local search in the parameter space with policy gradient $\nabla_\theta J(\pi_\theta) = \mathbb{E}_{\pi_\theta}\big[\sum_{t=0}^\infty A^{\pi_\theta}(s_t,a_t) \nabla_\theta \log \pi_\theta(a_t|s_t)\big]$ with the incremental update $\theta_{\text{new}} \leftarrow \theta + \alpha \nabla_\theta J(\pi_\theta)$ with some learning rate $\alpha > 0$. Alternatively, we can motivate the above gradient update with a trust region formulation. In particular, consider the following constrained optimization problem
\begin{align}
\max_{\theta_{\text{new}}} \mathbb{E}_{\pi_\theta}\big[&\frac{\pi_{\theta_\text{new}}(a_t|s_t)}{\pi_\theta(a_t|s_t)} A^{\pi_\theta}(s_t,a_t)\big], \nonumber \\
&||\theta_{\text{new}} - \theta||_2 \leq \epsilon,
\label{eq:trustregionpg}
\end{align}
for some $\epsilon > 0$. If we do a linear approximation of the objective in (\ref{eq:trustregionpg}), $\mathbb{E}_{\pi_\theta}\big[\frac{\pi_{\theta_\text{new}}(a_t|s_t)}{\pi_\theta(a_t|s_t)} A^{\pi_\theta}(s_t,a_t)\big] \approx \mathbb{E}_{\pi_\theta} \big[A^{\pi_\theta}(s_t,a_t)\big] + \nabla_\theta J(\pi_\theta)^T (\theta_{\text{new}} - \theta)$, we recover the gradient update by properly choosing $\epsilon$ given $\alpha$.

In such vanilla policy gradient updates, the training can suffer from occasionally large step sizes and never recover from a bad policy \citep{schulman2015}. The following variants are introduced to entail more stable updates.

\subsubsection{Trust Region Policy Optimization (TRPO)}
Trust Region Policy Optimization (TRPO) \citet{schulman2015} apply an information constraint on $\theta_{\text{new}}$ and $\theta$ to better capture the geometry on the parameter space induced by the underlying policy distributions, consider the following trust region formulation
\begin{align}
\max_{\theta_{\text{new}}} \mathbb{E}_{\pi_\theta}\big[&\frac{\pi_{\theta_\text{new}}(a_t|s_t)}{\pi_\theta(a_t|s_t)} A^{\pi_\theta}(s_t,a_t)\big], \nonumber \\
&\mathbb{E}_{\pi_\theta}\big[\mathbb{KL}[\pi_\theta || \pi_{\theta_{\text{new}}}]\big] \leq \epsilon.
\label{eq:trpo}
\end{align}
The trust region enforced by the KL divergence entails that the update according to (\ref{eq:trpo}) optimizes a lower bound of $J(\pi_\theta)$, so as to avoid accidentally taking large steps that irreversibly degrade the policy performance during training as in vanilla policy gradient (\ref{eq:trustregionpg}) \citep{schulman2015}. As a practical algorithm, TRPO further approximates the KL constraint in (\ref{eq:trpo}) by a quadratic constraint and approximately invert the constraint matrix by conjugate gradient iterations \citep{wright1999numerical}.

\paragraph{Actor Critic using Kronecker-Factored Trust Region (ACKTR)} As a more scalable and stable alternate to TRPO, ACKTR \citet{wu2017scalable} propose to invert the constraint matrix by K-Fac \citep{martens2015optimizing} instead of conjugate gradient iteration in TRPO. 

\subsubsection{Proximal Policy Optimization (PPO)}
For a practical algorithm, TRPO requires approximately inverting the Fisher matrices by conjugate gradient iterations. Proximal Policy Optimization (PPO) \citet{schulman2017} propose to approximate a trust-region method by clipping the likelihood ratios $\rho_t = \frac{\pi_{\theta_\text{new}}(a_t|s_t)}{\pi_\theta(a_t|s_t)}$ as $\bar{\rho}_t = \text{clip}(\rho_t,1-\eta,1+\eta )$ where $\text{clip}(x,a,b)$ clips the argument $x$ between $a$ and $b$. Consider the following objective
\begin{align}
\max_{\theta_{\text{new}}} \mathbb{E}_{\pi_\theta}\big[\min\{\rho_t & A^{\pi_\theta}(s_t,a_t),\bar{\rho}_t A^{\pi_\theta}(s_t,a_t)\}\big], \nonumber \\
&||\theta_{\text{new}} - \theta||_2 \leq \epsilon.
\label{eq:ppo}
\end{align}
 The intuition behind the objective (\ref{eq:ppo}) is when $A^{\pi_\theta}(s_t,a_t) > 0$, the clipping removes the incentives for the ratio $\rho_t$ to go above $1+\eta$, with similar effects for the case when $A^{\pi_\theta}(s_t,a_t) < 0$. PPO achieves more stable empirical performance than TRPO and involves only relatively cheap first-order optimization.

\subsection{Related Work}
\paragraph{On-policy Optimization.} Vanilla policy gradient updates are typically unstable \citep{schulman2015}. Natural policy \citet{kakade2002natural} apply information metric instead of Euclidean metric of parameters for updates. To further stabilize the training for deep RL, TRPO \citet{schulman2015} place an explicit KL divergence constraint between consecutive policies \citep{kakade2002approximately} and approximately enforces the constraints with conjugate gradients. PPO \citet{schulman2017} replace the KL divergence constraint by a clipping in the likelihood ratio of consecutive policies, which allows for fast first-order updates and achieves state-of-the-art performance for on-policy optimizations. On top of TRPO, ACKTR \citet{wu2017scalable} further improves the natural gradient computation with Kronecker-factored approximation \citep{george2018fast}. Orthogonal to the algorithmic advances, we demonstrate that policy optimizations with discrete/ordinal policies achieve consistent and significant improvements on all the above benchmark algorithms over baseline distributions.

\paragraph{Policy Classes.} Orthogonal to the algorithmic procedures for policy updates, one is free to choose any policy classes. In discrete action space, the only choice is a categorical distribution (or a discrete distribution). In continuous action space, the default baseline is factorized Gaussian \citep{schulman2015,schulman2017}. Gaussian mixtures, implicit generative models or even Normalizing flows \citep{rezende2015} can be used for more expressive and flexible policy classes \citep{tang2018implicit,tuomas2017,tuomas2018,tuomas2018b}, which achieve performance gains primarily for off-policy learning. One issue with aformentioned prior works is that they do not disentangle algorithms from distributions, it is therefore unclear whether the benefits result from a better algorithm or an expressive policy. To make the contributions clear, we make no changes to the on-policy algorithms and show the net effect of how the policy classes improve the performance. Motivated by the fact that unbounded distributions can generate infeasible actions, \citet{chou2017improving} propose to use Beta distribution and also show improvement on TRPO. Early prior work \citet{sharifflunar} also propose truncated Gaussian distribution but such idea is not tested on deep RL tasks. Complement to prior works, we propose discrete/ordinal policy as simple yet powerful alternates to baseline policy classes.

\paragraph{Discrete and Continuous Action Space.} Prior works have exploited the connection between discrete and continuous action space. For example, to solve discrete control tasks, \citet{van2009using,dulac2015deep} leverage the continuity in the underlying continuous action space for generalization across discrete actions. Prior works have also converted continuous control problems into discrete ones, e.g. \citet{pazis2009binary} convert low-dimensional control problems into discrete problems with binary actions. Surprisingly, few prior works have considered a discrete policy and apply off-the-shelf policy optimization algorithms directly. Recently, \citet{openai2018learning,jaskowski2018reinforcement} apply discrete policies to challenging hand manipulation and humanoid walking respectively. As a more comprehensive study, we carry out a full evaluation of discrete/ordinal policy on continuous control benchmarks and validate their performance gains. 

To overcome the explosion of action space, \citet{metz2018discrete} overcome the explosion by sequence prediction, but so far their strategy is only shown effective on relatively low-dimensional problems (e.g. HalfCheetah). \citet{tavakoli2018action} propose to avoid taking $\arg\max$ across all actions in Q-learning, by applying $\arg\max$ independently across dimensions. Their method is also only tested on a very limited number of tasks. As an alternative, we consider distributions that factorize across dimensions and we show that this simple technique yields consistent performance gains.

\paragraph{Ordinal Architecture.} When discrete variables have an internal ordering, it is beneficial to account for such ordering when modeling the categorical distributions. In statistics, such problems are tackled as ordinal regression or classification \citep{winship1984regression,chu2005gaussian,chu2007support}. Few prior works aim to combine ideas of ordinal regression with neural networks. Though \citet{cheng2008neural} propose to introduce ordering as part of the loss function, they do not introduce a proper probabilistic model and need additional techniques during inference. More recently, \citet{khan2012stick} motivate the \emph{stick-breaking} parameterization, a proper probabilistic model which does not introduce additional parameters compared to the original categorical distribution. In our original derivation, we motivate the architecture of \citep{khan2012stick} by transforming the loss function of \citep{cheng2008neural}. We also show that such additional inductive bias greatly boosts the performance for PPO/TRPO.

%In contrast to on-policy optimizations, discretizing continuous action space in off-policy setting is less obvious. Off-policy learning typically keeps track of a parametric action value function \citep{mnih2013} and leverages Bellman equations for updates. In continuous action space, the intractable $\arg\max$ operator is approximated by either a trained policy \citep{timothy2016} or the action value function is assumed to have structures that entail easy computations of $\arg\max$ \citep{gu2016continuous}. When the actions are discrete, it is not straightforward how to carry out the operator over an exponentially many combinations of atomic actions. Recently, \citep{metz2018discrete} proposes to model discrete action selection by sequence prediction.

\section{Discretizing Action Space for Continuous Control}
Without loss of generality, we assume the action space $\mathcal{A} = [-1,1]^m$. We discretize each dimension of the action space into $K$ equally spaced atomic actions. The set of atomic action for any dimension $i$ is $\mathcal{A}_i = \{\frac{2j}{K-1}-1\}_{j=0}^{K-1}$. To overcome the curse of dimensionality, we represent the distribution as factorized across dimensions. In particular, in state $s$, we specify a categorical distribution $\pi_{\theta_j}(a_j|s)$ over actions $a_j \in \mathcal{A}_j$ for each dimension $j$, with $\theta_j$ as parameters for this marginal distribution. Then we define the joint discrete policy $\pi(a|s) \coloneqq \Pi_{j=1}^m \pi_{\theta_j}(a_j|s)$ where $a = [a_0,a_1,...a_{K-1}]^T$. The factorization allows us to maintain a tractable distribution over joint actions, making it easy to do both sampling and training.

\subsection{Network Architecture}
The discrete policy is parameterized as follows. As in prior works \citep{schulman2015trust,schulman2017proximal}, the policy $\pi_\theta$ is a neural network that takes state $s$ as input, through multiple layers of transformation it will encode the state into a hidden vector $h(s) = f_\theta(s)$. For the $j$th action in the $i$th dimension of the action space, we output a logit $L_{ij} = w_{ij}^T h(s) + b_{ij} \in \mathbb{R}$ with parameters $w_{ij},b_{ij}$. For any dimension $i$, the $K$ logits $L_{ij},1\leq j \leq K$ are combined by soft-max to compute the probability of choosing action $j$, $p_{ij} = \text{softmax}(L_{ij}) (\coloneqq  \frac{\exp(L_{ij})}{\sum_{j=0}^{K-1} L_{ij}})$. By construction, the network has a fixed-size low-level parameter $\theta$, while the output layer has parameters $w_{ij},b_{ij}$ whose size scales linearly with $K$. 

%Other policy alternatives only differ in the output layer, while they all share the same architecture for the low-level encoding $h(s) = f_\theta(s)$. For example, a Gaussian policy will introduce parameters $w,b$ such that $\mu(s) = w^T h(s) + b$ represents the mean of the action distribution $a \sim \mathcal{N}(\mu(s),\sigma^2)$ where $\sigma$ is a separate standard deviation parameter.

\subsection{Understanding Discrete Policy for Continuous Control}

Here we briefly analyze the empirical properties of the discrete policy.

\begin{figure}[t]
\centering
\subfigure[\textbf{Bandit: Density}]{\includegraphics[width=.45\linewidth]{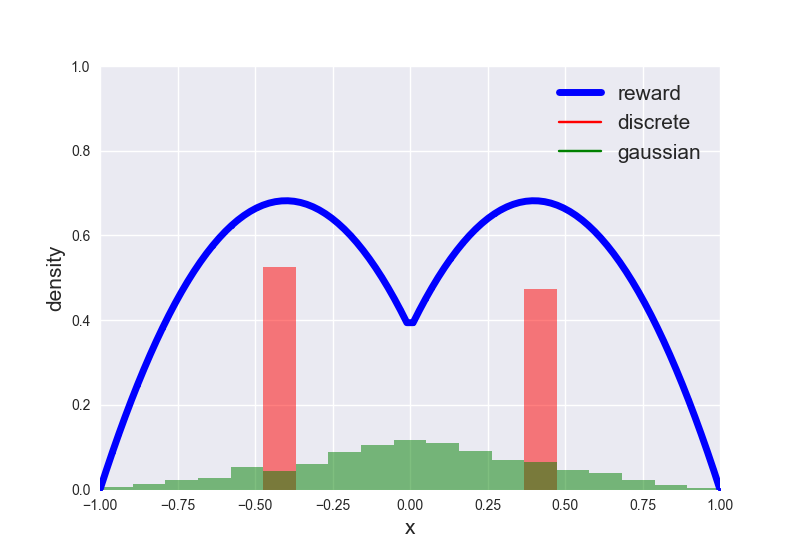}}
\subfigure[\textbf{Bandit: curves}]{\includegraphics[width=.45\linewidth]{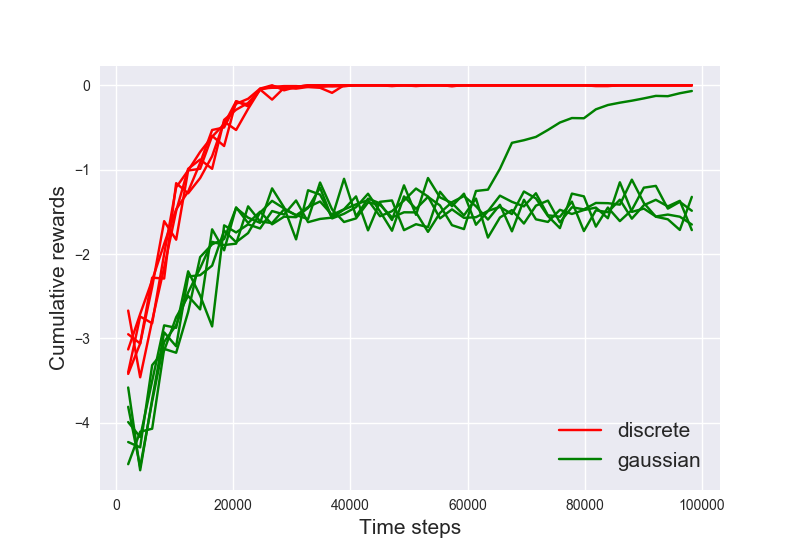}}
%\subfigure[\textbf{Gradient Variance}]{\includegraphics[width=.45\linewidth]{graph/variance}}
%\subfigure[\textbf{Control capacity}]{\includegraphics[width=.45\linewidth]{graph/bins}}
%\subfigure[Inverted Pendulum]{\includegraphics[width=.23\linewidth]{graph/benchmark_ppo_invertedpendulum}}
\caption{\small{Analyzing discrete policy: (a) Bandit example: comparison of normalized reward (blue), trained discrete policy density (red) and trained Gaussian policy density (green). (b) Bandit example: learning curves of discrete policy vs. Gaussian policy, we show five random seeds. Most seeds for Gaussian policy get stuck at a suboptimal policy displayed in (a) and all discrete policies reach a bi-modal optimal policy as in (a).}}
%(c) Variance of policy gradients on multiple tasks upon initialization, and comparison with the curve suggested by the simplified theoretical analysis above. The horizontal axis is the number of bins $K$ while the vertical axis is normalized such that the value is $1.0$ when $K=50$. The variance saturates quickly as $K \geq 10$. (d) Control capacity as a function of number of bins $K$. When $K$ is small, the control capacity is small and the policy cannot achieve good performance. When $K$ is large, the control capacity increases but training becomes harder, which leads to a potential drop in performance. 
\label{figure:control}
\end{figure}

\paragraph{Discrete Policy is more expressive than Gaussian.} Though discrete policy is limited on taking atomic actions, in practice it can represent much more flexible distributions than Gaussian when there are sufficient number of atomic actions (e.g. $K \geq 11$). Intuitively, discrete policy can represent multi-modal action distribution while Gaussian is by design unimodal. We illustrate this practical difference by a bandit example in Figure \ref{figure:control}. Consider a one-step bandit problem with $\mathcal{A} = [-1,1]$. The reward function for action $a$ is $r(a)$ illustrated as the Figure \ref{figure:control} (a) blue curve. We train a discrete policy with $K = 11$ and a Gaussian policy on the environment for $10^5$ steps and show their training curves in (b), with five different random seeds per policy. We see that 4 out 5 five Gaussian policies are stuck at a suboptimal policy while all discrete policies achieve the optimal rewards. Figure \ref{figure:control} (a) illustrates the density of a trained discrete policy (red) and a suboptimal Gaussian policy (green). The trained discrete policy is bi-modal and automatically captures the bi-modality of the reward function (notice that we did not add entropy regularization to encourage high entropy). The only Gaussian policy that achieves optimal rewards in (b) captures only one mode of the reward function.

For general high-dimensional problems, the reward landscape becomes much more complex. However, this simple example illustrates that the discrete policy can potentially capture the multi-modality of the landscape and achieve better exploration \citep{tuomas2017} to bypass suboptimal policies.

%\subsubsection{Analyzing the gradients for logit parameters}
%We now analyze the policy gradients of logit parameters. Models with $K$ atomic actions per dimension have a total of $Km$ ($m$ is the number of dimension of the action space) logits.

%!TEX encoding = UTF-8 UnicodeFor simplicity, we analyze linear models where logits are linear functions of states $L_{ij} = w_{ij}^T s + b_{ij}$. Consider the same one-step bandit problem as above, where the instant reward for action $a$ is $r(a)$. The $i$th discretized action is $a_i,0\leq i\leq K-1$. Here in one dimension the parameters are $\{w_i,b_i\}_{i=0}^{K-1}$. The policy gradient estimator

%\paragraph{Training Costs and Control Capacity.}
\paragraph{Effects of the Number of Atomic Actions $K$.}
Choosing a proper number of atomic actions per dimension $K$ is critical for learning. The trade-off comes in many aspects: \textbf{(a)} Control capacity. When $K$ is small the discretization is too coarse and the policy does not have enough capacity to achieve good performance. 
\textbf{(b)} Training difficulty. When $K$ increases, the variance of policy gradients also increases. We detail the analysis of policy gradient variance in Appendix B. We also present the combined effects of \text{(a)} and \text{(b)} in Appendix B, where we find that the best performance is obtained when $7\leq K  \leq 15$, and setting $K$ either too large or too small will degrade the performance. \textbf{(c)} Model parameters and computational Costs. Both the number of model parameters increases linearly and computational costs grow linearly in $K$. We present detailed computational results in Appendix B. 

%\begin{minipage}{\linewidth}
%\centering
%\captionof{table}{MuJoCo Benchmark Tasks} \label{tab:title} 
%\begin{tabular}{ C{0.8in}  *4{C{.6in}} }\toprule[1.5pt]
%\bf Action size & \bf $K=5$ & \bf $K=11$ & \bf $K=30$ & \bf $K=100$ \\\midrule
%\bf Percentage   & 116\% & 120\% & 143\% & 240\%  \\
%\bottomrule[1.25pt]
%\end {tabular}\par
%\bigskip
%\small{Table 1: Computational costs measured in wall time on Reacher task. PPO with Gaussian policy is normalized to be 100\% and we report the normalized time for discrete policy. Each number is averaged over 3 random seeds. The increase in costs is roughly linear in $K$ but can be more severe when the action dimension increases.}
%\end{minipage}

\section{Discrete Policy with Ordinal Architecture}
\subsection{Motivation}
When the continuous action space is discretized, we treat continuous variables as discrete and discard important information about the underlying continuous space. It is therefore desirable to incorporate the notion of continuity when paramterizing distributions over discrete actions.

%For discrete variables with internal ordering, it is valuable to account for the underlying continuity. Prior works have combined statistical machine learning techniques with ordinal regression \citep{chu2005gaussian,chu2007support}, but little work combines ordinal regression with neural networks \citep{cheng2008neural}. Though \citet{cheng2008neural} proposes an architecture and loss function that heuristically models continuity between predicted classes, it does not produce a proper probabilistic model. Following the intuition introduced in \citep{cheng2008neural}, we propose a feed-forward architecture that parameterizes a discrete distribution while accounting for the internal ordering between classes through the architecture's inductive bias.

\subsection{Ordinal Distribution Network Architecture}
For simplicity, we discuss the discrete distribution over only one action dimension. Recall previously that a typical feed-forward architecture that produces discrete distribution over $K$ classes produces $K$ logits $L_i$ at the last layer and derives the probability via a softmax $p_i = \text{softmax}(L_i), 1\leq i\leq K$. In the ordinal architecture, we retain these logits $L_i$ and first transform them via a sigmoid function $s_i = \sigma(L_i)$. Then we compute the final logits as 
\begin{align}
    L^\prime_i = \sum_{j\leq i} \log s_j + \sum_{j>i} \log (1-s_j),\forall 1\leq i\leq K,
    \label{eq:ordinal}
\end{align}
and derive the final output probability via a softmax $p^\prime_i = \text{softmax}(L^\prime_i)$. The actions are sampled according to this discrete distribution $a_i \sim p_i^\prime$.

This architecture is very similar to the \emph{stick-breaking} parameterization introduced in \cite{khan2012stick}, where they argue that such parameterization is beneficial when the samples drawn from class $k$ can be easily separated from samples drawn from all the classes $j>k$. In our original derivation, we motivate the ordinal architecture from the loss function of \citep{cheng2008neural} and we show that the intuition behind (\ref{eq:ordinal}) is more clear from this perspective. We show the intuition below with a $K$-way classification problem, where the classes are internally ordered as $\{1,2,...K\}$.

\paragraph{Intuition behind  (\ref{eq:ordinal}).} For clarity, let $\mathbf{x},\mathbf{y} \in \mathbb{R}^K$ such that $0\leq x_i,y_i \leq 1$ and define the  stable cross entropy  $\text{CE}(\mathbf{x},\mathbf{y}) \coloneqq -\sum_{i=1}^K x_i \log \max\{y_i,\epsilon\})$ with a very small $\epsilon>0$ to avoid numerical singularity. For a sample from class $k$, the $K-$way classification loss based on (\ref{eq:ordinal}) is $-L_k^\prime = \text{CE}(\mathbf{t}_k, \mathbf{s})$, where the predicted vector $\mathbf{s} = [s_1,s_2...s_K]$ and a target vector $\mathbf{t}_k = [1,1...0]$ with first $k$ entries to be $1$s and others $0$s. The intuition becomes clear when we interpret $\mathbf{t}_k$ as a continuous encoding of the class $k$ (instead of the one-hot vector) and $\mathbf{s}$ as a intermediate vector from which we draw the final prediction. We see that the continuity between classes is introduced through the loss function, for example $\text{CE}(\mathbf{t}_k,\mathbf{t}_{k+1}) < \text{CE}(\mathbf{t}_k,\mathbf{t}_{k+2})$, i.e. the discrepancy between class $k$ and $k+1$ is strictly smaller than that between $k$ and $k+2$. On the contrary, such information cannot be introduced by one-hot encoding: let $\mathbf{e}_{k}$ be the one-hot vector for class $k$, we always have e.g. $\text{CE}(\mathbf{e}_k,\mathbf{e}_{k+1}) =  \text{CE}(\mathbf{e}_k,\mathbf{e}_{k+2})$, i.e. we introduce no discrepancy relationship between classes. While \citet{cheng2008neural} introduce such continuous encoding techniques, they do not propose proper probabilistic models and require additional techniques at inference time to make predictions. Here, the ordinal architecture (\ref{eq:ordinal}) defines a proper probabilistic model that implicitly introduces internal ordering between classes through the parameterization, while maintaining all the probabilistic properties of discrete distributions.

In summary, the oridinal architecture (\ref{eq:ordinal}) introduces additional dependencies between logits $L_i$ which implicitly inject the information about the class ordering. In practice, we find that this generally brings significant performance gains during policy optimization.

%The intuition behind (\ref{eq:ordinal}) is more clear when compared to the original categorical parameterization: when the $k^\as$ action has positive advantage $A^\pi(s,k^\ast) > 0$, the policy gradient will increase $\pi(k^\ast|s)$. The original parameterization achieves this by increasing the $k^\ast$th logit $L_{k^\ast}$. This implies very little about the logits of other actions (except for decreasing all other logits since $\sum_k \text{softmax}(L_k) = 1$.) For the ordinal parameterization, this is achieved by increasing $s_i, i\leq k^\ast$ and decreasing $s_i, s>k^\ast$ in (\ref{eq:ordinal}). Such updates (very vaguely) imply that all actions $i\leq k^\ast$ have positive advantages and $i > k^\ast$ have negative advantages. Leveraging 

%We discuss both the intuition behind (\ref{eq:ordinal}) and its connection to \citep{cheng2008neural}. 

%So far we have empirically through experiments that such ordinal architecture significantly improves upon its categorical baseline architecture on PPO. The improvement on other algorithms (TRPO and ACKTR) is not as obvious. \textcolor{red}{explain?}

%\begin{wrapfigure}{R}{.4\textwidth}
%\includegraphics[width=.38\textwidth]{graph/bins}
%\caption{\small{Control capacity}}
%\label{figure:control}
%\end{wrapfigure}

\section{Experiments}
Our experiments aim to address the following questions: \textbf{(a)} Does discrete policy improve the performance of baseline algorithms on benchmark continuous control tasks? \textbf{(b)} Does the ordinal architecture further improve upon discrete policy? \textbf{(c)} How sensitive is the performance to hyper-parameters, particularly to the number of bins per action dimension? 

For clarity, we henceforth refer to \emph{discrete policy} as with the discrete distribution, and \emph{ordinal policy} as with the ordinal architecture. To address \textbf{(a)}, we carry out comparisons in two parts: (1) We compare discrete policy (with varying $K$) with Gaussian policy over baseline algorithms (PPO, TRPO and ACKTR), evaluated on benchmark tasks in gym MuJoCo \citep{brockman2016,todorov2008}, rllab \citep{duanxi2016}, roboschool \citep{schulman2015} and Box2D. Here we pay special attention to Gaussian policy because it is the default policy class implemented in popular code bases \citep{baselines}; (2) We compare with other architectural alternatives, either straihghtforward architectural variants or those suggested in prior works \citep{chou2017improving}. We evaluate their performance on high-dimensional tasks with complex dynamics (e.g. Walker, Ant and Humanoid). All the above results are reported in Section 5.1. To address \textbf{(b)}, we compare discrete policy with ordinal policy with PPO in Section 5.2 (results for TRPO are also in Section 5.1). To address \text{(c)}, we randomly sample hyper-parameters for Gaussian policy and discrete policy and compare their quantiles plots in Section 5.3 and Appendix C. 

%\footnote{As noted in \textcolor{red}{forgot the ref!}, HalfCheetah \citep{brockman2016} is not a task with complex dynamics. Here complex tasks refer to Walker, Ant and Humanoids.}. All results are reported in Section 5.1. To address \textbf{(b)}, we compare discrete policy with ordinal policy with PPO in Section 5.2. To address \text{(c)}, we randomly sample hyper-parameters for both policies and compare their quantiles plots in Section 5.3. 

\paragraph{Implementation Details.} As we aim to study the net effect of the discrete/ordinal policy with on-policy optimization algorithms, we make minimal modification to the original PPO/TRPO/and ACKTR algorithms originally implemented in OpenAI baselines \citep{baselines}. We leave all hyper-parameter settings in Appendix A. 

\begin{figure*}[t]
\centering
%\subfigure[Reacher]{\includegraphics[width=.23\linewidth]{graph/benchmark_ppo_reacher}}
%\subfigure[Swimmer]{\includegraphics[width=.23\linewidth]{graph/benchmark_ppo_swimmer}}
%\subfigure[Inverted Pendulum]{\includegraphics[width=.23\linewidth]{graph/benchmark_ppo_invertedpendulum}}
%\subfigure[Double Pendulum]{\includegraphics[width=.23\linewidth]{graph/benchmark_ppo_inverteddoublependulum}}
%\subfigure[\textbf{Hopper}]{\includegraphics[width=.23\linewidth]{graph/benchmark_ppo_hopper}}
\subfigure[\textbf{HalfCheetah + PPO}]{\includegraphics[width=.23\linewidth]{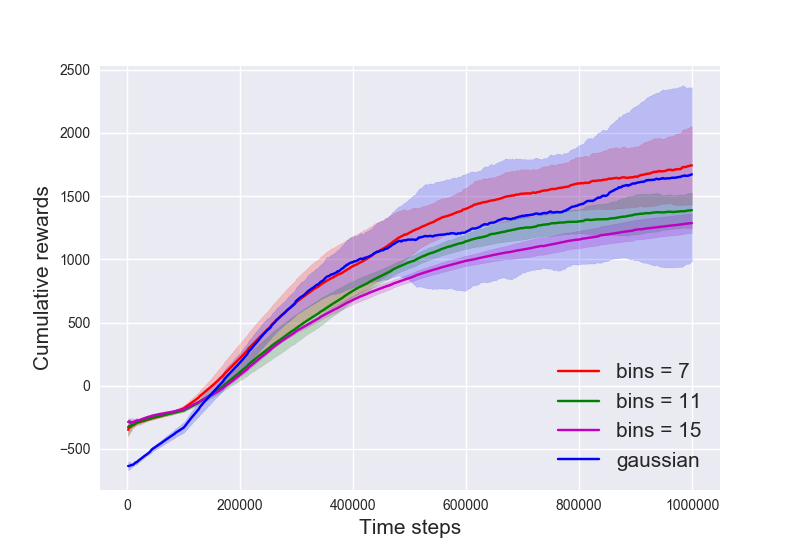}}
\subfigure[\textbf{Ant  + PPO}]{\includegraphics[width=.23\linewidth]{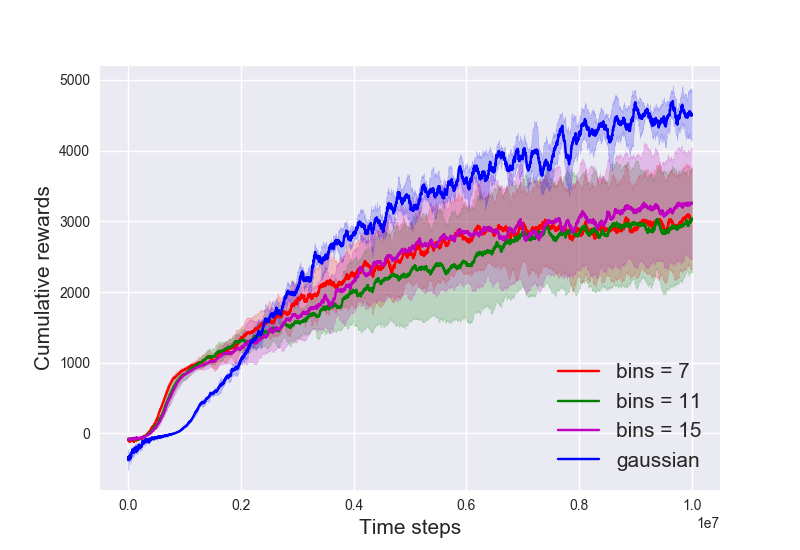}}
\subfigure[\textbf{Walker + PPO}]{\includegraphics[width=.23\linewidth]{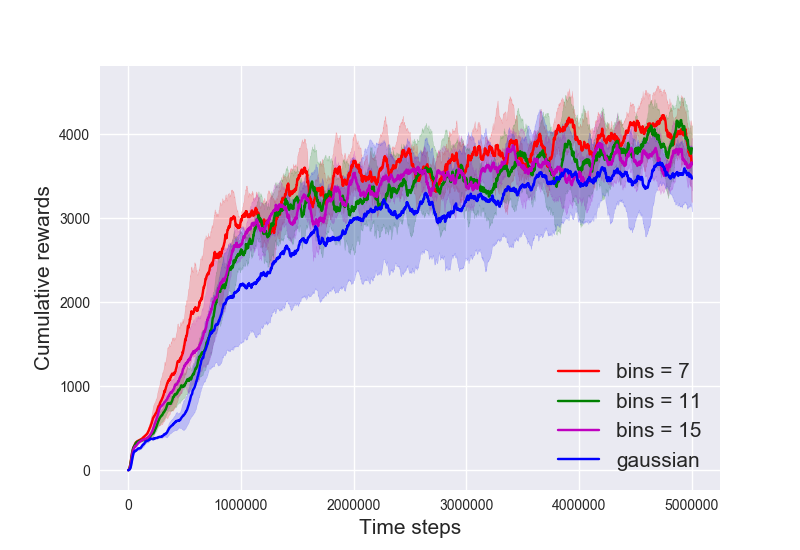}}
\subfigure[\textbf{Humanoid (R) + PPO} ]{\includegraphics[width=.23\linewidth]{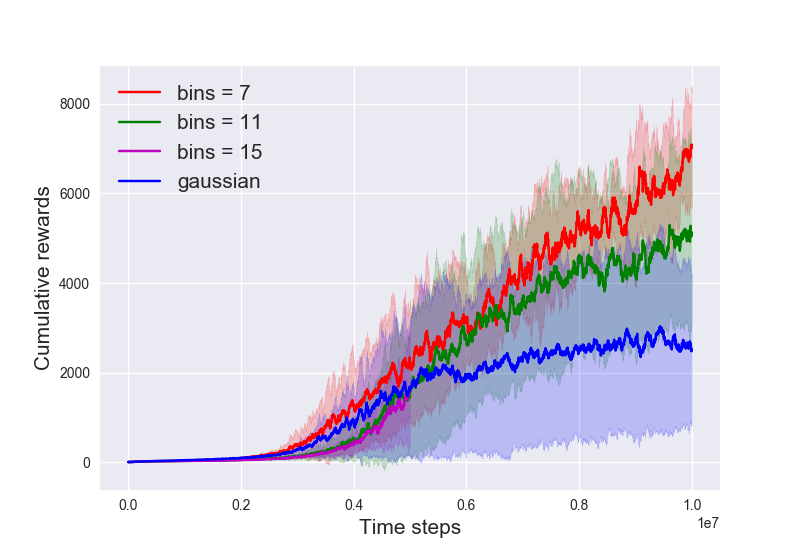}}
\subfigure[\textbf{Sim. Human. (R) + PPO}]{\includegraphics[width=.23\linewidth]{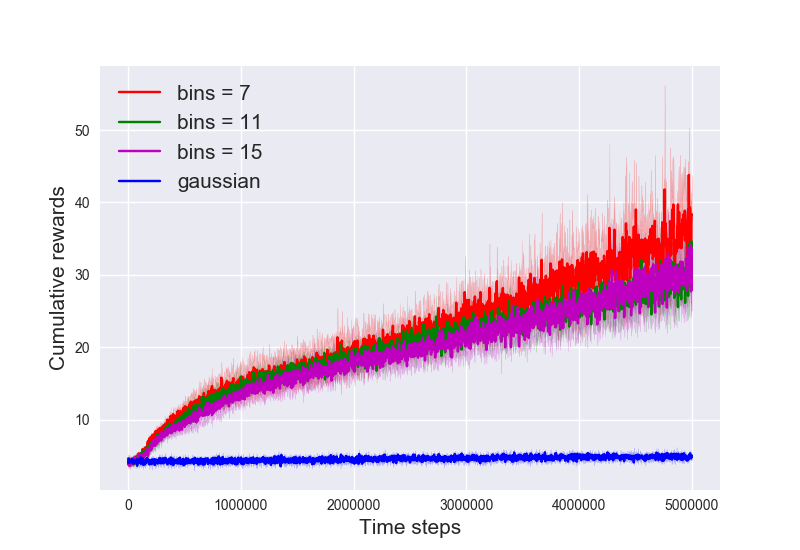}}
\subfigure[\textbf{Humanoid + PPO}]{\includegraphics[width=.23\linewidth]{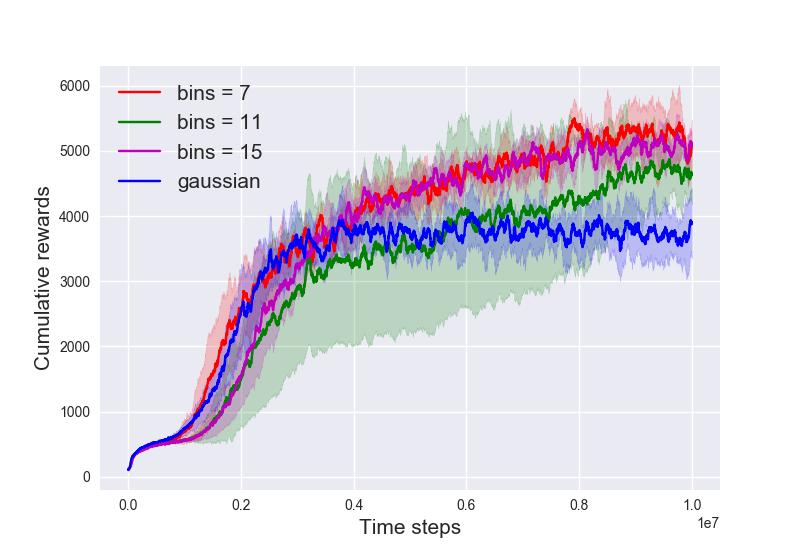}}
\subfigure[\textbf{HalfCheetah + TRPO}]{\includegraphics[width=.23\linewidth]{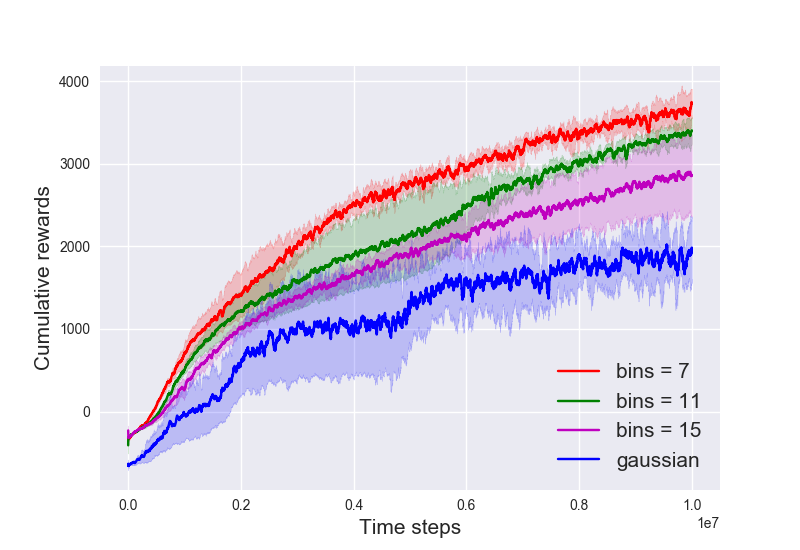}}
\subfigure[\textbf{Ant + TRPO}]{\includegraphics[width=.23\linewidth]{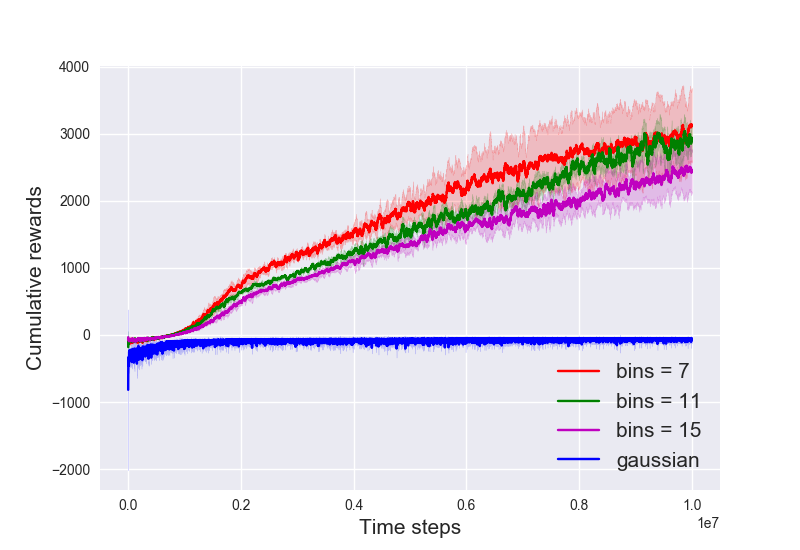}}
\subfigure[\textbf{Walker + TRPO}]{\includegraphics[width=.23\linewidth]{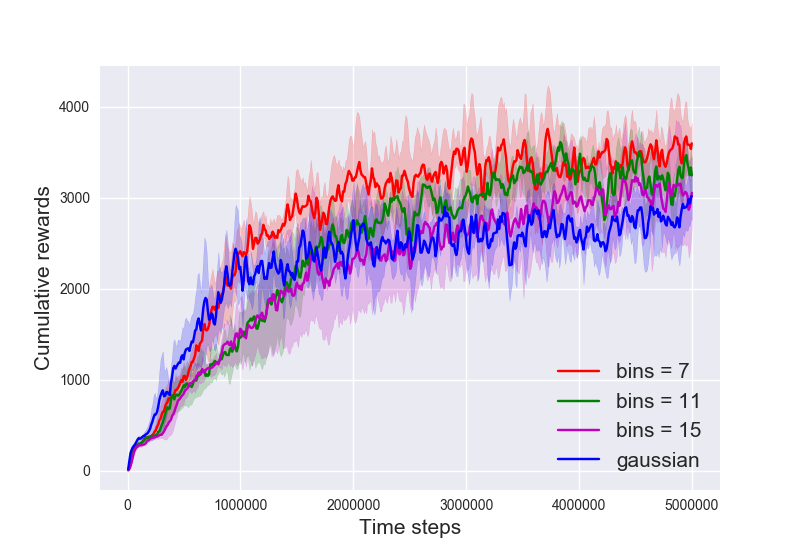}}
\subfigure[\textbf{Sim. Human. (R) + TRPO} ]{\includegraphics[width=.23\linewidth]{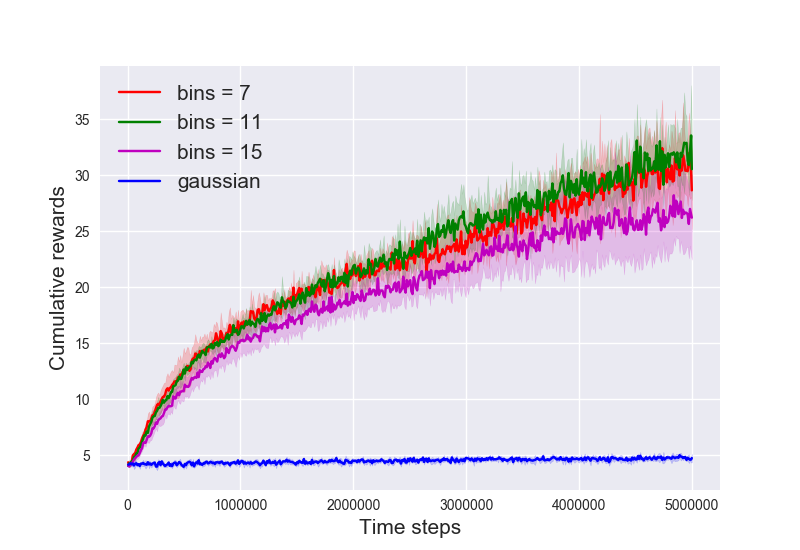}}
\subfigure[\textbf{Humanoid (R) + TRPO} ]{\includegraphics[width=.23\linewidth]{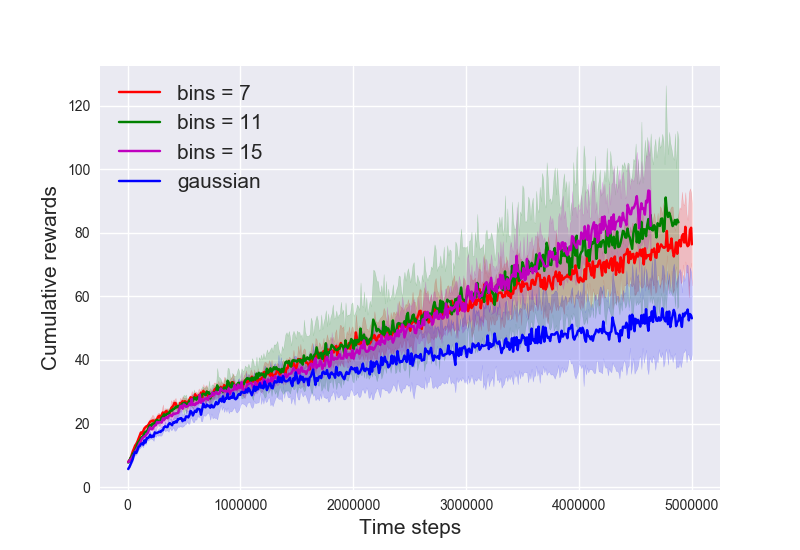}}
\subfigure[\textbf{Humanoid + TRPO}]{\includegraphics[width=.23\linewidth]{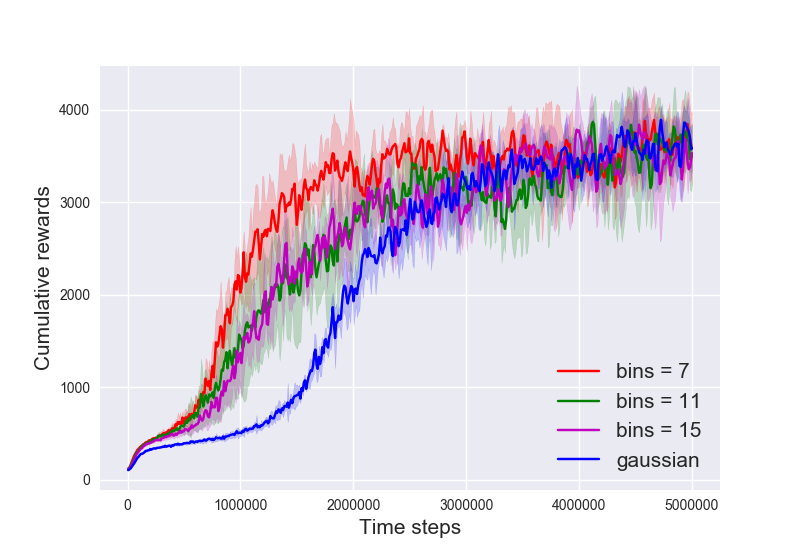}}
%\subfigure[\textbf{HumanoidStandup}]{\includegraphics[width=.23\linewidth]{graph/benchmark_ppo_humanoidstandup}}
\caption{\small{MuJoCo Benchmarks: learning curves of PPO on OpenAI gym MuJoCo locomotion tasks. Each curve is averaged over 5 random seeds and shows $\text{mean} \pm \text{std}$ performance. Each curve corresponds to a different policy architecture (Gaussian or discrete actions with varying number of bins $K=7,11,15$). Vertical axis is the cumulative rewards and horizontal axis is the number of time steps. Discrete actions significantly outperform Gaussian on Humanoid tasks. Tasks with (R) are from rllab.}}
\label{figure:benchmarkppo}
\end{figure*}

\subsection{Benchmark performance}

\begin{figure}[h]
\centering
\subfigure[\textbf{Humanoid + PPO}]{\includegraphics[width=.45\linewidth]{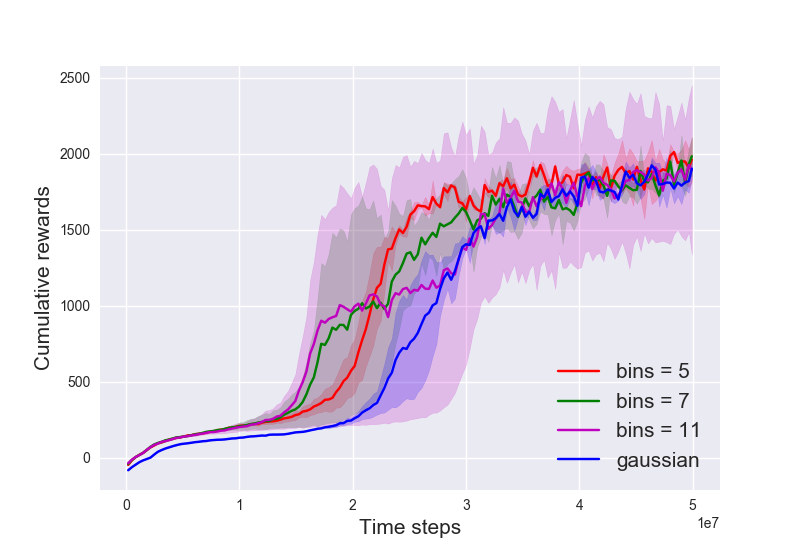}}
\subfigure[\textbf{Humanoid + TRPO}]{\includegraphics[width=.45\linewidth]{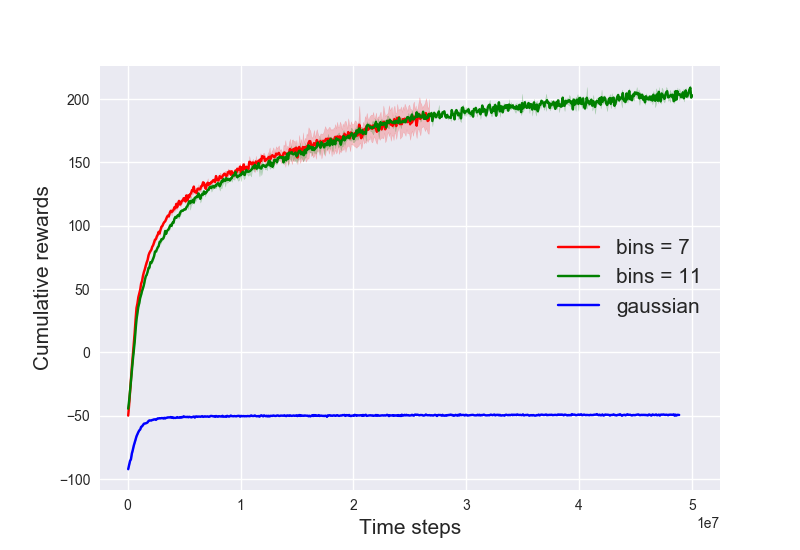}}
\subfigure[\textbf{Flagrun + PPO}]{\includegraphics[width=.45\linewidth]{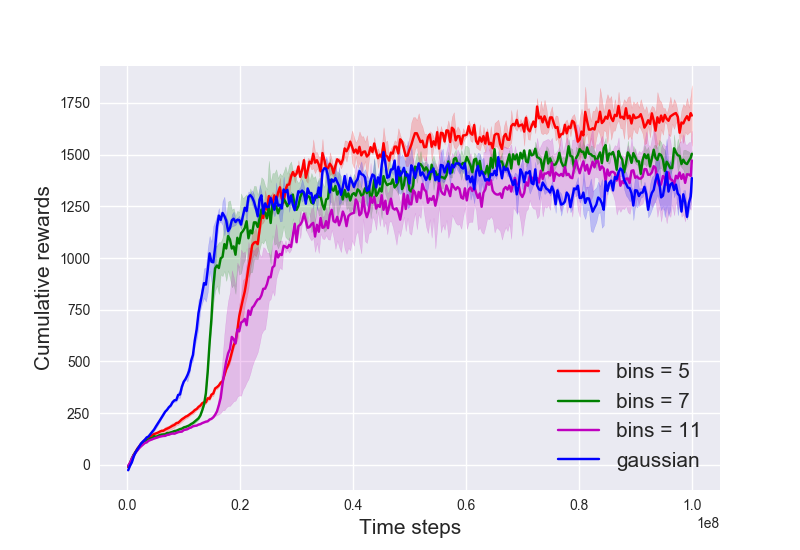}}
\subfigure[\textbf{Flagrun + TRPO}]{\includegraphics[width=.45\linewidth]{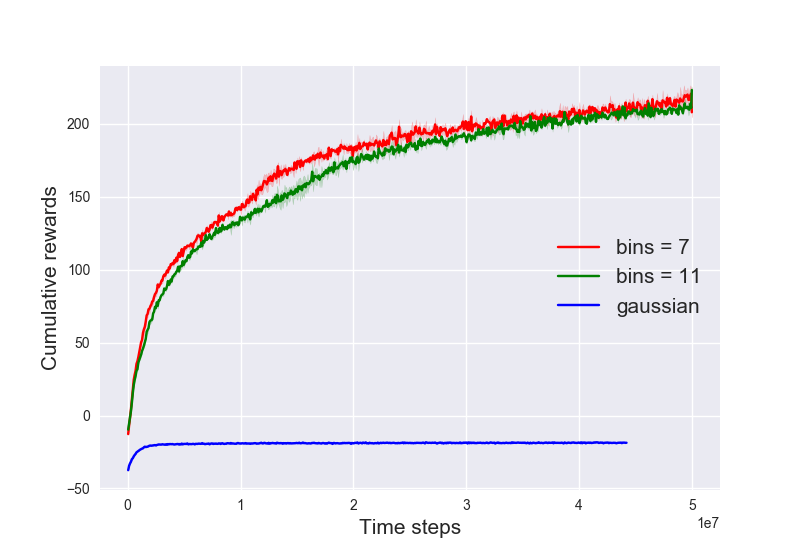}}
\subfigure[\textbf{FlagrunHarder + PPO}]{\includegraphics[width=.45\linewidth]{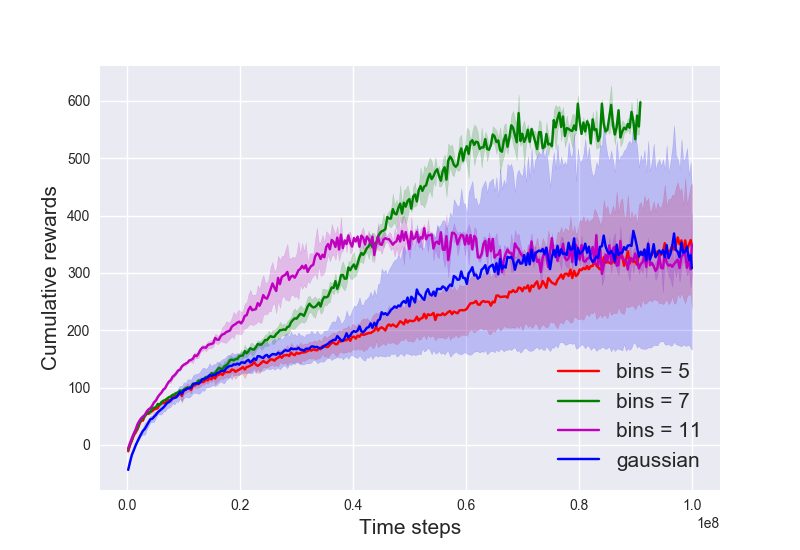}}
\subfigure[\textbf{FlagrunHarder + TRPO}]{\includegraphics[width=.45\linewidth]{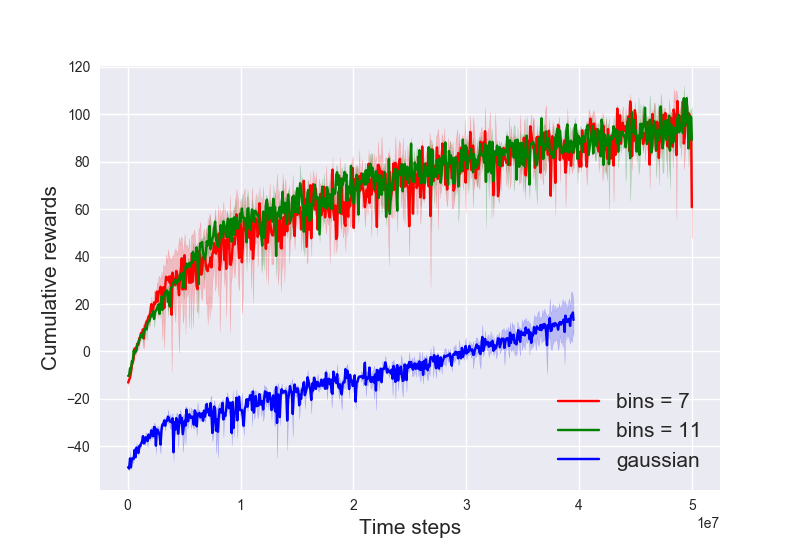}}
\caption{\small{Roboschool Humanoid Benchmarks: learning curves of PPO/TRPO on Roboschool Humanoid locomotion taskse. Each curve corresponds to a different policy architecture (Gaussian or discrete actions with varying number of bins $K=5,7,11$). Discrete policies outperform Gaussian policy on all Humanoid tasks and the performance gains are more significant with TRPO.}}
\label{figure:benchmarkroboschool}
\end{figure}

\begin{figure}[t]
\centering
%\subfigure[Reacher]{\includegraphics[width=.23\linewidth]{graph/benchmark_ppo_reacher}}
%\subfigure[Swimmer]{\includegraphics[width=.23\linewidth]{graph/benchmark_ppo_swimmer}}
%\subfigure[Inverted Pendulum]{\includegraphics[width=.23\linewidth]{graph/benchmark_ppo_invertedpendulum}}
%\subfigure[Double Pendulum]{\includegraphics[width=.23\linewidth]{graph/benchmark_ppo_inverteddoublependulum}}
\subfigure[\textbf{Walker}]{\includegraphics[width=.45\linewidth]{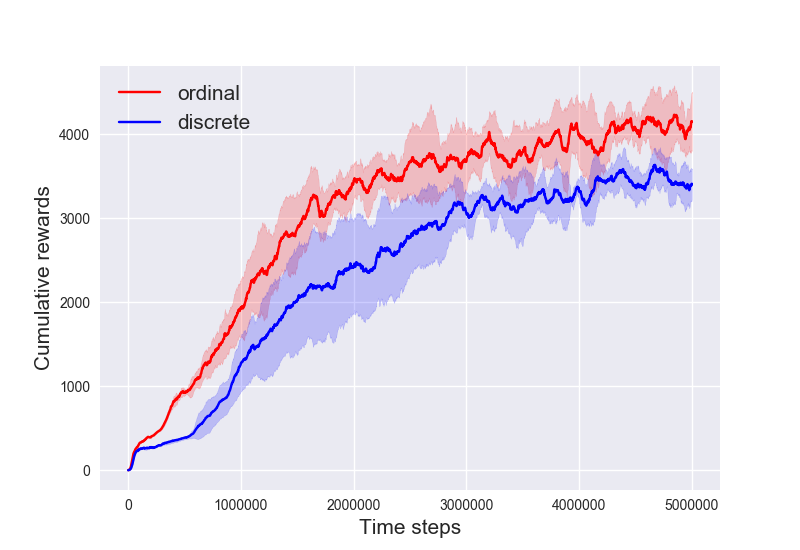}}
%\subfigure[\textbf{HalfCheetah}]{\includegraphics[width=.23\linewidth]{graph/ordinal_discrete_halfcheetah.png}}
\subfigure[\textbf{Ant}]{\includegraphics[width=.45\linewidth]{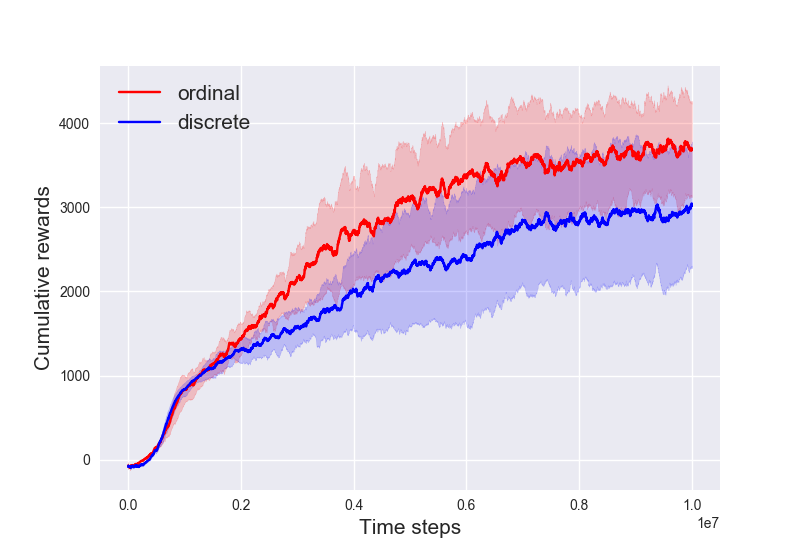}}
\subfigure[\textbf{Humanoid (R)}]{\includegraphics[width=.45\linewidth]{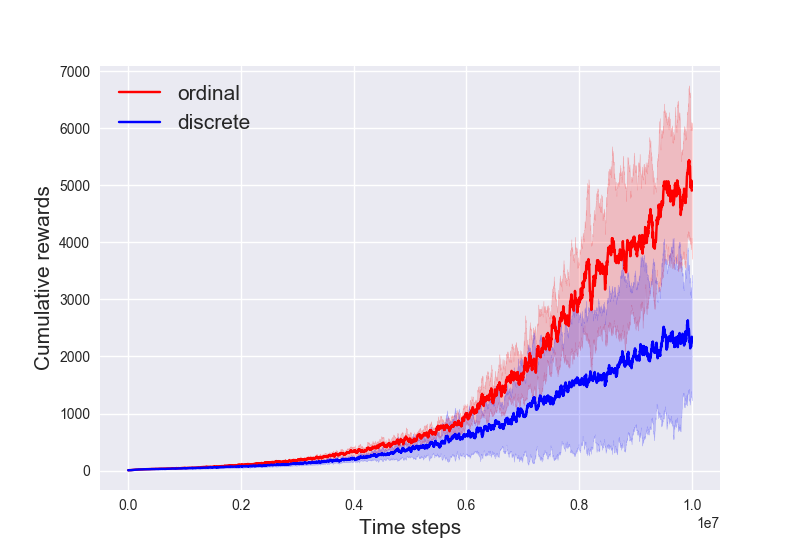}}
\subfigure[\textbf{Sim. Humanoid (R)}]{\includegraphics[width=.45\linewidth]{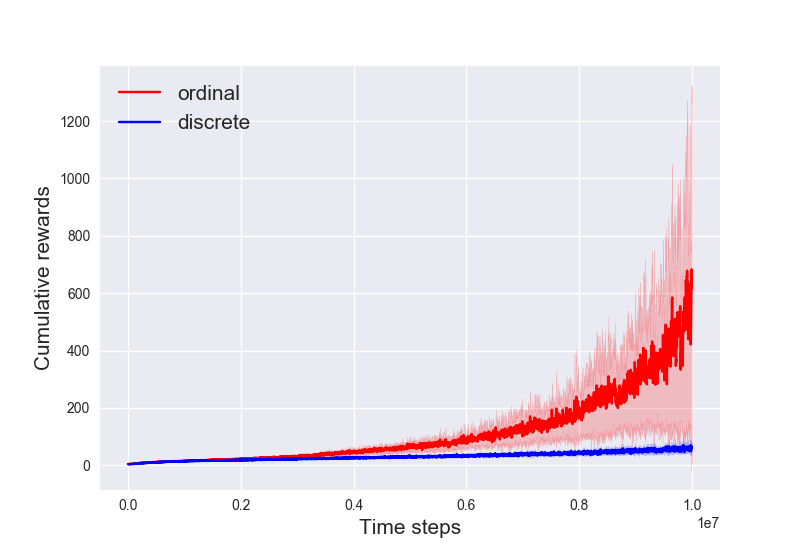}}
\subfigure[\textbf{Humanoid}]{\includegraphics[width=.45\linewidth]{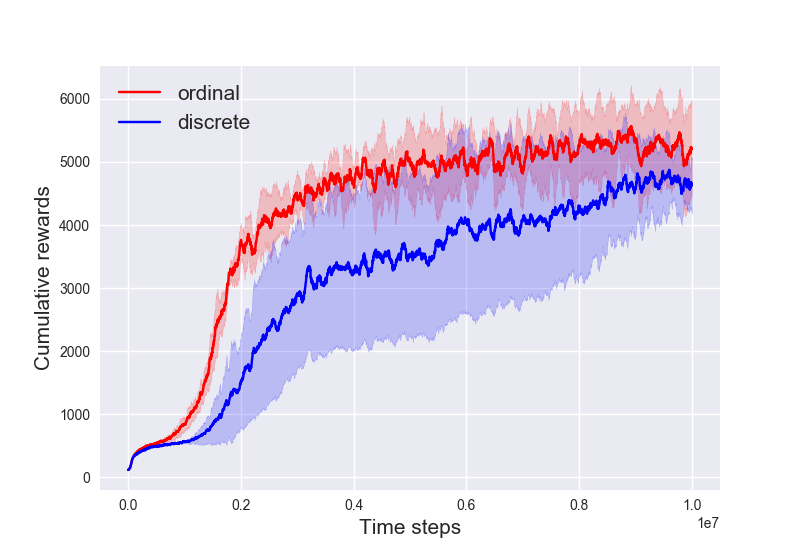}}
\subfigure[\textbf{Humanoid Standup}]{\includegraphics[width=.45\linewidth]{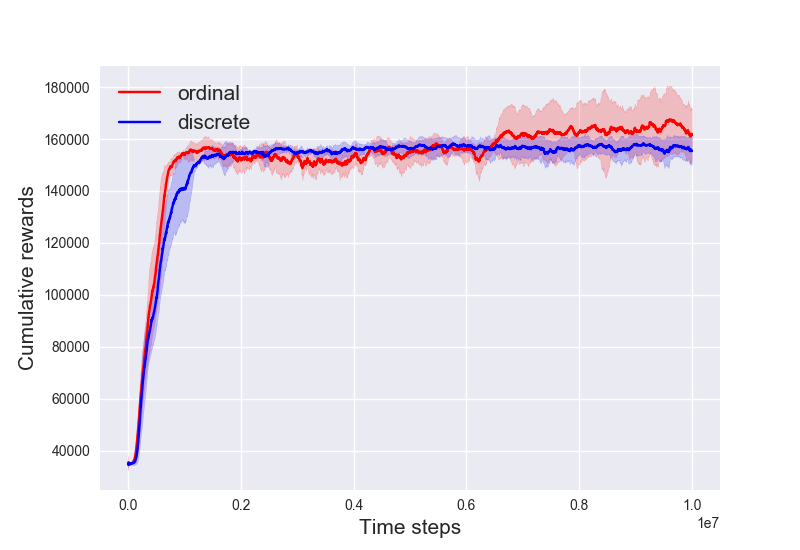}}
\caption{\small{MuJoCo Benchmarks: learning curves of PPO + discrete policy vs. PPO + ordinal policy on OpenAI gym MuJoCo locomotion tasks. All policies have $K=11$. We see that for each task, ordinal policy outperforms discrete policy.}}
\label{figure:benchmarkordinal}
\end{figure}

All benchmark comparison results are presented in plots (Figure \ref{figure:benchmarkppo},\ref{figure:benchmarkroboschool}) or tables (Table \ref{table:comparewithotherdists},\ref{table:fullcomparison}). For plots, we show the learning curves of different policy classes trained for a fixed number of time steps. The x-axis shows the time steps while the y-axis shows the cumulative rewards. Each curve shows the average performance with standard deviation shown in shaded areas. Results in Figure \ref{figure:benchmarkppo},\ref{figure:benchmarkordinal} are averaged over 5 random seeds and Figure \ref{figure:benchmarkroboschool} over 2 random seeds. In Table \ref{table:comparewithotherdists},\ref{table:fullcomparison} we train all policies for a fixed number of time steps and we show the average $\pm$ standard deviation of the cumulative rewards obtained in the last 10 training iterations. 

\paragraph{PPO/TRPO - Comparison with Gaussian Baselines.}
We evaluate PPO/TRPO with Gaussian against PPO/TRPO with discrete policy on the full suite of MuJoCo control tasks and display all results in Figure \ref{figure:benchmarkppo}. For PPO, on tasks with relatively simple dynamics, discrete policy does not necessarily enjoy significant advantages over Gaussian policy. For example, the rate of learning of discrete policy is comparable to Gaussian for HalfCheetah (Figure \ref{figure:benchmarkppo}(a)) and even slightly lower on Ant \ref{figure:benchmarkppo}(b)). However, on high-dimensional tasks with very complex dynamics (e.g. Humanoid, Figure \ref{figure:benchmarkppo}(d)-(f)), discrete policy significantly outperforms Gaussian policy. For TRPO, the performance gains by discrete policy are also very consistent and significant.

We also evaluate the algorithms on Roboschool Humanoid tasks as shown in Figure \ref{figure:benchmarkroboschool}. We see that discrete policy achieves  better results than Gaussian across all tasks and both algorithms. The performance gains are most significant with TRPO (Figure \ref{figure:benchmarkroboschool}(b)(d)(f)), where we see Gaussian policy barely makes progress during training while discrete policy has very stable learning curves. For completeness, we also evaluate PPO/TRPO with discrete policy vs. Gaussian policy on Box2D tasks and see that the performance gains are significant. Due to space limit, We present Box2D results in Appendix C. 

By construction, when discrete policy and Gaussian policy have the same encoding architecture $f_\theta(s)$ shown in Section 3, discrete policy has many more parameters than Gaussian policy. A critical question is whether we can achieve performance gains by simply increasing the number of parameters? We show that when we train a Gaussian policy with many more parameters (e.g. $128$ hidden units per layer), the policy does not perform as well. This validates our speculation that the performance gains result from a more carefully designed distribution class rather than larger networks.

\paragraph{PPO - Comparison with Off-Policy Baselines.}
To further illustrate the strength of PPO with discrete policy on high-dimensional tasks with very complex dynamics, we compare PPO with discrete policy against state-of-the-art off-policy algorithms on Humanoid tasks (Humanoid-v1 and Humanoid rllab) \footnote{Humanoid-v1 has $|\mathcal{S}| = 376, |\mathcal{A}| = 17$ and Humanoid rllab has $|\mathcal{S}| = 142, |\mathcal{A}| = 21$. Both tasks have very high-dimensional observation space and action space.}. Such algorithms include DDPG \citep{timothy2016}, SQL \citep{tuomas2017}, SAC \citep{tuomas2018} and TD3 \citep{fujimoto2018addressing}, among which SAC and TD3 are known to achieve significantly better performance on MuJoCo benchmark tasks over other algorithms. Off-policy algorithms reuse samples and can potentially achieve orders of magnitude better sample efficiency than on-policy algorithms. For example, it has been commonly observed in prior works \citep{tuomas2018,tuomas2018b,fujimoto2018addressing} that SAC/TD3 can achieve state-of-the-art performance on most benchmark control tasks for only $10^6$ steps of training, on condition that off-policy samples are heavily replayed. In general, on-policy algorithms cannot match such level of fast convergence because samples are quickly discarded. However, for highly complex tasks such as Humanoid even off-policy algorithms take many more samples to learn, potentially because off-policy learning becomes more unstable and off-policy samples are less informative. In Table \ref{table:comparewithotherdists}, we record the  
final performance of off-policy algorithms directly from the figures in \citep{tuomas2018} following the practice of \citep{mania2018simple}. The final performance of PPO algorithms are computed as the average $\pm$ std of the returns in the last 10 training iterations across 5 random seeds. All algorithms are trained for $10^7$ steps. We observe in Table \ref{table:comparewithotherdists} that PPO + discrete (ordinal) actions achieve comparable or even better results than off-policy baselines. This shows that for general complex applications, PPO $+$ discrete/ordinal is still as competitive as the state-of-the-art off-policy methods.

\paragraph{PPO/TRPO - Comparison with Alternative Architectures.} We also compare with straightforward architectural alternatives: Gaussian with $\text{tanh}$ non-linearity as the output layer, and Beta distribution \citep{chou2017improving}. The primary motivation for these architectures is that they naturally bound the sampled actions to the feasible range ($[-1,1]$ for gym tasks). By construction, our proposed discrete/ordinal policy also bound the sampled actions within the feasible range. In Table \ref{table:fullcomparison}, we show results for PPO/TRPO where we select the best result from $K \in \{7,11,15\}$ for discrete/ordinal policy. We make several observations from results in Table \ref{table:fullcomparison}: (1) Bounding actions (or action means) to feasible range does not consistently bring performance gains, because we observe that Gaussian $+\ \text{tanh}$ and Beta distribution do not consistently outperform Gaussian. This is potentially because the parameterizations that bound the actions (or action means) also introduce challenges for optimization. For example, Gaussian $+\ \text{tanh}$ bounds the action means $\mu_\theta(s) \in [-1,1]$, this implies that in order for $\mu_\theta(s) \approx \pm 1$ the parameter $\theta$ must reach extreme values, which is hard to achieve using SGD based methods. (2) Discrete/Ordinal policy  achieve significantly better results consistently across most tasks. Combining (1) and (2), we argue that the performance gains of discrete/ordinal policies are due to reasons beyond a bounded action distribution.

%In particular, when learning rate is large for PPO (trust region size is large for TRPO), Beta policy can terminate prematurely due to numerical errors. The training can be completed with smaller learning rate for PPO  (smaller trust region size for TRPO) but the average performance can greatly degrade. \textcolor{red}{a fair way to report beta} In general, we observe that discrete/ordinal policy achieve consistently and significantly better results than other distributions. This shows that discrete/ordinal policy achieve better optimization for reasons beyond bounding actions to remove implicit bias (which is the main motivation behind Beta distribution) while being much more numerically stable.

Here we discuss the results for Beta distribution. In our implementation we find training with Beta distribution tends to generate numerical errors when the 
update is more aggressive (e.g. PPO learning rate $3\cdot10^{-5}$ or TRPO trust region size is $0.01$). More conservative updates (e.g. e.g. PPO learning rate $3\cdot10^{-6}$ or TRPO trust region size is $0.001$) reduce numerical errors but also greatly degrade the learning performance. We suspect that this is because the Beta distribution parameterization (Appendix A and \citep{chou2017improving}) is numerically unstable and we discuss the potential reason in Appendix A.  In Table \ref{table:fullcomparison}, the results for Beta distribution is recorded as the performance of the last 10 iterations before the training terminates (potentially prematurely due to numerical errors). The potential advantages of Beta distribution are largely offset by the unstable training. We show more results in Appendix C.

\begin{table*}[t]
  \caption{\small{A comparison of PPO with discrete/ordinal policy with state-of-the-art baseline algorithms on Humanoid benchmark tasks from OpenAI gym and rllab. For each task, we show the average rewards achieved after training the agent for a fixed number of time steps. The results for PPO + discrete/ordinal/Gaussian policy are mean performance averaged over 5 random seeds (see Figure \ref{figure:benchmarkppo}. The results for DDPG, SQL, SAC and TD3 are approximated based on the figures in \citep{tuomas2018}. The results for PPO is consistent with results in \citep{tuomas2018}. Even compared to off-policy algorithms, PPO + ordinal policy achieves state of the art performance across both tasks.}}
\vskip 0.15in
\begin{center}
\begin{small}
\begin{sc}
\begin{tabular}{lccccccc}
\toprule
 \bf Tasks  & \bf DDPG & \bf SQL & \bf SAC & \bf TD3& \bf PPO + Gaussian & \bf PPO + discrete  & \bf PPO + ordinal \\\midrule
Humaoid-v1        & $\approx 500$ & $\approx 5500$ & $\mathbf{\approx 6000}$ &$\mathbf{\approx 6000}$& $\approx 4000$ &  $5119 \pm 151$ & $\mathbf{6018 \pm 239}$ \\
Humanoid(rllab) & $<500$ & $\approx 2000$ & $\mathbf{\approx 5500}$ &$< 500$&  $\approx 2500$ & $4084 \pm 1312$ & $\mathbf{4884 \pm 1562}$ \\ 
\bottomrule
\end{tabular}
\end{sc}
\end{small}
\end{center}
\vskip -0.1in
\label{table:comparewithotherdists}
\end{table*}
%beta humanoid two seeds$3050 \pm 2480$

\begin{table*}[t]
\caption{Comparison across a range of policy alternatives (Gaussian, Gaussian $+ \text{tanh}$, and Beta distribution \citep{chou2017improving}). All policies are optimized with \textbf{PPO/TRPO}. All tasks are training for 10M steps. Results are the average $\pm$ std performance for the last 10 training iterations.  Top two results (with highest average) are highlighted in bold font. Tasks with (R) are from rllab.}
\label{policy}
\vskip 0.15in
\begin{center}
\begin{small}
\begin{sc}
\begin{tabular}{lccccc}
\toprule
\bf PPO &  \bf Gaussian & \bf Gaussian+$\text{tanh}$ & \bf Beta  & \bf Discrete & \bf Ordinal  \\
\midrule
Walker2d & $3500 \pm 360$ & $3274 \pm 251$ & $274 \pm 6$  & $\mathbf{3390 \pm 190}$ & $\mathbf{4249 \pm 239}$   \\
Ant & $\mathbf{4445 \pm 194}$ & $\mathbf{4622 \pm 171}$ & $3112 \pm 173$  & $3256 \pm 778$ & $3690 \pm 557$   \\
HalfCheetah & $1598 \pm 23$ & $1566 \pm 26$ & $1193 \pm 24$  & $\mathbf{4824\pm 199}$ & $\mathbf{3477 \pm 1497}$   \\
Humanoid & $3905 \pm 502$ & $4007 \pm 698$ & $2680 \pm 2493$  & $\mathbf{5119 \pm 151}$ & $\mathbf{6018 \pm 403}$   \\
HumanoidStandup & $\mathbf{166446 \pm 18348}$ & $160983 \pm 3842$ & $155362 \pm  8657$  & $161618 \pm  10224$ & $\mathbf{170275 \pm 19316}$   \\
Humanoid (R) & $2522 \pm 1684$ & $5863 \pm 1288$ & $2680 \pm 2493$ & $\mathbf{4084 \pm 1312}$ & $\mathbf{4884 \pm 1562}$ \\
Sim. Humanoid (R) & $5.1 \pm  0.4$ & $4.3 \pm 0.5$ & $4.4 \pm 0.6$  & $\mathbf{214\pm 136}$ & $\mathbf{801 \pm  569}$\\
\midrule
\bf TRPO &  \bf  & \bf  & \bf   & \bf & \bf   \\
\midrule
%Walker2d & $?$ & $3?$ & $?$  & $\mathbf{3390 \pm 190}$ & $\mathbf{4249 \pm 239}$   \\
Ant  &  $-76 \pm 14$  & $-89 \pm 13$ & $2362 \pm 305$  & $\mathbf{2687 \pm  556}$ & $\mathbf{2977 \pm 266}$   \\
HalfCheetah  &  $1576 \pm 782$  & $386 \pm 78$ & $1643 \pm 819$   & $\mathbf{3081 \pm 766}$ & $\mathbf{3352 \pm 1196}$   \\
Humanoid & $1156 \pm 163$ & $\mathbf{6350 \pm 486}$ & $3812 \pm 1973$  & $\mathbf{3908 \pm 117}$ & $3577 \pm 272$   \\
Humanoid Standup  &  $137955 \pm 9238$  & $133558 \pm 9238$ & $111497 \pm 15031$ & $\mathbf{142640 \pm 2343}$ & $\mathbf{143418 \pm 8638}$   \\
Humanoid (R) & $65 \pm 8$ & $38 \pm 2$ & $38 \pm 3$ & $\mathbf{84 \pm 24}$ & $\mathbf{161 \pm 26}$ \\
Sim. Humanoid (R) & $6.5 \pm 0.2$  &  $4.4 \pm 0.1$  & $4.2 \pm 0.2$  & $\mathbf{42 \pm 6}$ & $\mathbf{93 \pm  28}$\\
\bottomrule
\end{tabular}
\end{sc}
\end{small}
\end{center}
\vskip -0.1in
\label{table:fullcomparison}
\end{table*}

\paragraph{ACKTR - Comparison with Gaussian Baselines.} We show results for ACKTR in Appendix C. We observe that for tasks with complex dynamics, discrete policy still achieves performance gains over its Gaussian policy counterpart.

\subsection{Discrete Policy vs. Ordinal Policy}
 In Figure \ref{figure:benchmarkordinal}, we evaluate PPO $+$ discrete policy  and PPO $+$ ordinal policy on high-dimensional tasks. Across all presented tasks, ordinal policy achieves significantly better performance than  discrete policy both in terms of asymptotic performance and speed of convergence. Similar results are also presented in table \ref{table:fullcomparison} where we show that PPO $+$ ordinal policy achieves comparable performance as efficient off-policy algorithms on Humanoid tasks. We also compare these two architectures when trained with TRPO. The comparison of the trained policies can be found in table \ref{table:comparewithotherdists}. For most tasks, we find that ordinal policy still significantly improves upon discrete policy. 

Summarizing the results for PPO/TRPO, we conclude that the ordinal architecture introduces useful inductive bias that improves policy optimization. We note that \emph{sticky-breaking} parameterization (\ref{eq:ordinal}) is not the only parameterization that leverages natural orderings between discrete classes. We leave as promising future work how to better exploit task specific ordering between classes.

\subsection{Sensitivity to Hyper-parameters}
Here we evaluate the policy classes' sensitivity to more general hyper-parameters, such as learning rate $\alpha$, number of bins per dimension $K$ and random seeds. We present the results of PPO in Appendix C. For PPO with Gaussian, we uniformly sample $\log_{10} \alpha \in [-6.0,-3.0]$ and one of $5$ random seeds. For PPO with discrete actions, we further uniformly sample $K \in \{7,11,15\}$. For each benchmark task, we sample $30$ hyper-parameters and show the quantile plot of the final performance. As seen in Appendix C, PPO with discrete actions is generally more robust to such hyper-parameters than Gaussian.

%\begin{figure}[h]
%\centering
%\subfigure[Roboschool Reacher]{\includegraphics[width=.23\linewidth]{graph/hyperparams_trpo_roboschoolreacher}}
%\subfigure[Hopper]{\includegraphics[width=.23\linewidth]{graph/hyperparams_trpo_hopper}}
%\subfigure[HalfCheetah]{\includegraphics[width=.23\linewidth]{graph/hyperparams_ppo_halfcheetah}}
%\subfigure[Roboschool Ant]{\includegraphics[width=.23\linewidth]{graph/hyperparams_ppo_roboschoolant}}
%\subfigure[Walker]{\includegraphics[width=.23\linewidth]{graph/benchmark_acktr_walker2d}}
%\subfigure[Humanoid]{\includegraphics[width=.23\linewidth]{graph/hyperparams_ppo_humanoid}}
%\subfigure[SimpleHumanoid (rllab)]{\includegraphics[width=.23\linewidth]{graph/hyperparams_ppo_simplehumanoidrllab}}
%\subfigure[Humanoid]{\includegraphics[width=.23\linewidth]{graph/benchmark_acktr_humanoid}}
%\subfigure[HumanoidStandup]{\includegraphics[width=.23\linewidth]{graph/benchmark_ppo_humanoidstandup}}
%\caption{\small{MuJoCo Benchmark : learning curves on MuJoCo locomotion tasks. Each curve is averaged over 6 random seeds and shows $\text{mean} \pm \text{std}$ performance. Each curve corresponds to a different policy representation (Red: Implicit, Green: GMM $K=5$, Yellow: GMM $K=2$, Blue: Gaussian). Vertical axis is the cumulative rewards and horizontal axis is the number of time steps}}
%\label{figure:benchmarkdist}
%\end{figure}

\section{Conclusion}
We have carried out a systemic evaluation of action discretization for continuous control across baseline on-policy algorithms and baseline tasks. Though the idea is simple, we find that it greatly improves the performance of baseline algorithms, especially on high-dimensional tasks with complex dynamics. We also show that the ordinal architecture which encodes the natural ordering into the discrete distribution, can further boost the performance of baseline algorithms. 

\section{Acknowledgements}
This work was supported by an Amazon Research Award (2017).  The authors also acknowledge the computational resources provided by Amazon Web Services (AWS).

\bibliographystyle{apa}
\bibliography{your_bib_file.bib}

\newpage
\onecolumn

\appendix
\section{Hyper-parameters}
 All implementations of algorithms (PPO, TRPO, ACKTR) are based on OpenAI baselines \citep{baselines}. All environments are based on OpenAI gym \citep{brockman2016}, rllab \citep{duanxi2016} and Roboschool \citep{schulman2017}.

We present the details of each policy class as follows.

\paragraph{Gaussian Policy.} Factorized Gaussian Policies are represented as $\pi_\theta(\cdot|s) = \mathbb{N}(\mu_\theta(s),\Sigma)$ with $\mu_\theta(s)$ as a two-layer neural network with $64$ hidden units per layer for PPO and ACKTR and $32$ units per layer for TRPO. The covariance matrix $\Sigma$ is diagonal $\Sigma_{ii} = \sigma_i^2$, with each $\sigma_i$ a single variable shared across all states. Such hyper-parameter settings are default with baselines.

\paragraph{Discrete Policy.} Discrete policies are represented as $\pi_\theta(\cdot|s) = \Pi_{i=1}^m \pi_{\theta_i}(\cdot|s)$ with $\pi_{\theta_i}(\cdot|s)$ a categorical distribution over atomic actions. We specify $K$ atomic actions across each dimension, evenly spaced between $-1$ and $1$. In each action dimension, the categorical distribution is specified by a set of logits $l_j(s)$ ($l_j(s)$ for action $j$ in state $s$) and $l_j(s)$ is parameterized to be a neural network with the same architecture as the factorized Gaussian above.

\paragraph{Ordinal Policy.} Ordinal policies are augmented with an ordinal parameterization compared to the discrete policies. Ordinal policy has exactly the same number of parameters as the discrete policy.

\paragraph{Gaussian$+ \text{tanh}$ Policy.} The architecture is the same as above but the final layer is added a $\tanh$ transformation to ensure that the mean $\mu_\theta(s) \in [-1,1]$.

\paragraph{Beta Policy.} A Beta policy has the form $\pi(\alpha_\theta(s),\beta_\theta(s))$ where $\pi$ is a Beta distribution with parameters $\alpha_\theta(s),\beta_\theta(s)$. Here, $\alpha_\theta(s)$ and $\beta_\theta(s)$ are shape/rate parameters parameterized by two-layer neural network $f_\theta(s)$ with a softplus at the end, i.e. $\alpha_\theta(s) = \log(\exp(f_\theta(s)) + 1) + 1$, following \citep{chou2017improving}. Actions sampled from this distribution have a strictly finite support. We notice that this parameterization introduces potential instability during optimization: for example, when we want to converge on policies that sample actions at the boundary, we require $\alpha_\theta(s) \rightarrow \infty$ or $\beta_\theta(s) \rightarrow \infty$, which might be very unstable. We also observe such instability in practice: when the trust region size is large (e.g. $\epsilon = 0.01$) the training can easily terminate prematurely due to numerical errors. However, reducing the trust region size (e.g. $\epsilon = 0.001$) will stabilize the training but degrade the performance. The results for Beta policy in the main paper are obtained under trust region size $\epsilon = 0.01$ for TRPO and learning rate $3 \cdot 10^{-5}$ for PPO. These hyper-parameters are chosen such that the policy achieves fairly fast rate of learning (compared to other policy classes) at the cost of more numerical errors (which lead to premature termination).

\paragraph{Others Hyper-parameters.} Value functions are two-layer neural networks with $64$ hidden units per layer for PPO and ACKTR and $32$ hidden units per layer for TRPO. For PPO, the learning rate is tuned from $\{3\cdot 10^{-5}, 3\cdot 10^{-4}\}$. For TRPO, the KL constraint parameter $\epsilon$ is tuned from $\{0.01,0.001\}$. For ACKTR, the KL constraint parameter is tuned from $\{0.02,0.002\}$. All other hyper-parameters are default parameters in the baselines implementations.

\section{Effects of the Number of Atomic Actions}
\paragraph{Variance of Policy Gradients.} We analyze the variance of policy gradients when the continuous action space is discretized. For an easy analysis, we assume that the policy architecture $\pi_\theta$ is as follows: the policy $\pi_\theta$ is in general a neural network (or any differentiable functions) that takes state $s$ as input, through multiple layers of transformation it will encode the state into a hidden vector $h = f_\theta(s)$. For the $i$th dimension of the action space, for the $j$th action in $\mathcal{A}_i$, we output a logit $L_{ij} = w_{ij}^T h + b_{ij} \in \mathbb{R}$ by parameters $w_{ij},b_{ij}$. For any dimension $i$, $K$ logits are combined by soft-max to compute the probability of choosing action $j$, $p_{ij} = \text{softmax}(L_{ij}) (\coloneqq  \frac{\exp(L_{ij})}{\sum_{j=0}^{K-1} L_{ij}})$. As noted, the number of model parameters scale linearly with $K$. 

To compare the variance of policy gradients across models with varying $K$, we analyze the gradients of parameters $\theta$ that encode $s$ into $h$. Such parameters are shared by all models. For simplicity, we consider a one step bandit problem with action space $\mathcal{A} = [-1,1]$. The instant reward for action $a$ is $r(a) = R, \forall a$ where $R$ is a fixed constant. Since we have only one action dimension, let $p(j)$ be the probability of taking the $j$th action $0\leq j\leq K-1$. We also assume that upon initialization the policy has very high entropy $p(j) \approx \frac{1}{K}$. The policy gradient estimator is 
\begin{align}
\hat{g}_\theta = r(a_j) \nabla_\theta \log p_\theta(j), \ j\sim p_\theta(j) \nonumber
\end{align}
Under this setting, the policy gradient $\mathbb{E}\big[\hat{g}_\theta \big] = \nabla_\theta J(\pi_\theta) = 0$ and the variance is 
\begin{align}
\mathbb{V}\big[\hat{g}_\theta\big] &= \sum_{j=1}^K r(a_j) (\nabla_\theta \log p(j))^2 p(j) \nonumber \\&\approx R^2 \frac{1}{K} \sum_{j=1}^K (\nabla_\theta \log p(j))^2 \nonumber \\
&\approx R^2 \frac{1}{K} (\sum_{j=1}^K (\nabla_\theta L_j - \frac{1}{K}\sum_{k=0}^{K-1} \nabla_\theta L_k))^2, \nonumber
\label{eq:variance}\\
\end{align}
where approximations come from replacing $p(j) \approx \frac{1}{K}$. Notice that $\nabla_\theta L_j$ does not depend on $N$ since each logit $L_j$ has an independent dependency on $\theta$. Under conventional neural network initializations (all weight and bias matrices of $\theta$ and $w_{ij},b_j$ are independently initialized), $\nabla_\theta L_j,0\leq j\leq K-1$ are i.i.d. random variables with their randomness stemming from the random initialization of neural network parameters. Denote $\mathbb{E}_{\text{init}}\big[\cdot\big]$ as the expectation w.r.t. neural network initializations, we analyze the expectation of (\ref{eq:variance})
\begin{align}
\mathbb{E}_{\text{init}} \big[ \mathbb{V}\big[\hat{g}_\theta\big]\big] &\approx R^2\frac{K-1}{K} \sigma^2 \sim \frac{K-1}{K}, 
\end{align}
where $\sigma^2 = \mathbb{V}_{\text{init}}\big[\nabla_\theta L_j\big]$ is the variance of the logit gradients. 
%\begin{wrapfigure}{R}{.4\textwidth}
%\includegraphics[width=.38\textwidth]{graph/variance}
%\caption{\small{Variance}}
%\label{figure:variance}
%\end{wrapfigure}

\begin{figure}[t]
\centering
\subfigure[\textbf{Gradient Variance}]{\includegraphics[width=.45\linewidth]{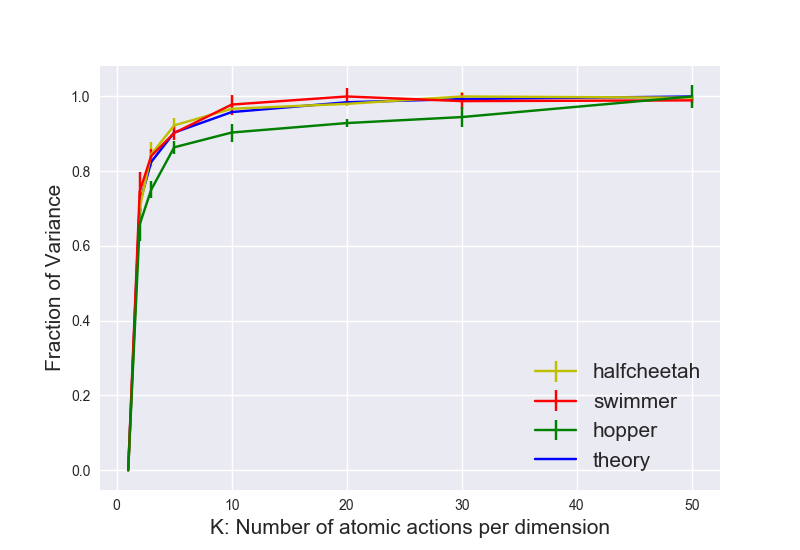}}
\subfigure[\textbf{Control capacity}]{\includegraphics[width=.45\linewidth]{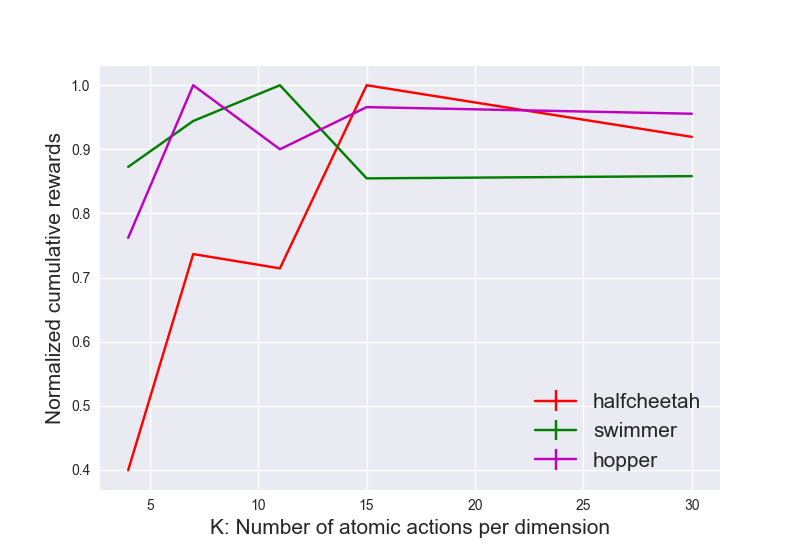}}
%\subfigure[Inverted Pendulum]{\includegraphics[width=.23\linewidth]{graph/benchmark_ppo_invertedpendulum}}
\caption{\small{Analyzing discrete policy: (a) Variance of policy gradients on multiple tasks upon initialization, and comparison with the curve suggested by the simplified theoretical analysis above. The horizontal axis is the number of bins $K$ while the vertical axis is normalized such that the value is $1.0$ when $K=50$. The variance saturates quickly as $K \geq 11$. (b) Control capacity as a function of number of bins $K$. When $K$ is small, the control capacity is small and the policy cannot achieve good performance. When $K$ is large, the control capacity increases but training becomes harder, which leads to a potential drop in performance. }}
\label{figure:variance}
\end{figure}

Though the above derivation makes multiple restrictive assumptions, we find that it also largely matches the results for more complex scenarios. For multiple MuJoCo tasks, we compute the empirical variance of policy gradients upon random initializations of network parameters. In Figure \ref{figure:variance}(a) we compare the empirical variance against the predicted variance. We normalize the variance such that the variance at $K=50$ is $1.0$. With the same hyper-parameters (including batch-size for each update), policy gradients have larger variance for models with more fined discretization (large $K$) and will be harder to optimize. 

\paragraph{Combined Effects of Control Capacity and Variance.}
When $K$ increases the policy has larger capacity for control. However, as analyzed above, the policy gradient variance also increases with $K$, which makes policy optimization more difficult using SGD based methods.

The combined effects can be observed in Figure \ref{figure:variance}(b). We train policies with various $K$ for a fixed number of time steps and evaluate their performance at the end of training. We find that the best performance is obtained when $7\leq K\leq 15$. When $K$ is small (e.g. $K=2$) the performance degrades drastically, due to the lack of control capacity. When $K$ is large (e.g. $K \approx 50$), the performance only slightly degrades: this might be because the variance of the policy gradient almost saturates when $K \geq 11$ as shown in Figure \ref{figure:variance}.

\paragraph{Model Parameters and Training Costs.}
Both the number of model parameters and training costs scale linearly with $K$. In Table \ref{table:traincost} we present the computational results for the training costs: we train discrete policies (on Reacher-v1) with various $K$ for a fixed number of time steps and record the wall time. The results are standardized such that Gaussian policy is $100\%$.

\begin{table}[t]
\caption{\small{Computational costs measured in wall time on Reacher task. PPO with Gaussian policy is normalized to be 100\% and we report the normalized time for discrete policy. Each number is averaged over 3 random seeds. The increase in costs is roughly linear in $K$ but can be more severe when the action dimension increases.}}
\vskip 0.15in
\begin{center}
\begin{small}
\begin{sc}
\begin{tabular}{lcccc}
\toprule
\bf Action size & \bf $K=5$ & \bf $K=11$ & \bf $K=30$ & \bf $K=100$ \\\midrule
\bf Percentage   & 116\% & 120\% & 143\% & 240\%  \\
\bottomrule
\end{tabular}
\end{sc}
\end{small}
\end{center}
\vskip -0.1in
\label{table:traincost}
\end{table}

\section{Additional Experiments}
\subsection{PPO}
\begin{figure}[h]
\centering
\subfigure[\textbf{Reacher}]{\includegraphics[width=.23\linewidth]{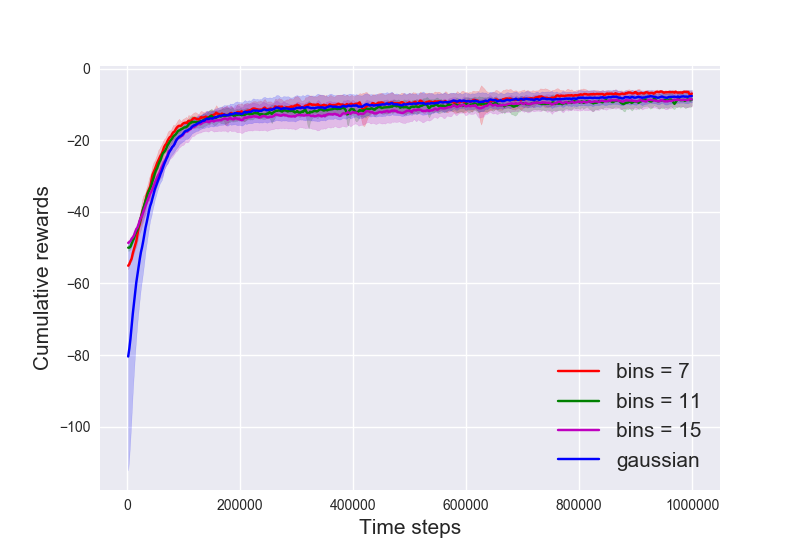}}
\subfigure[\textbf{Swimmer}]{\includegraphics[width=.23\linewidth]{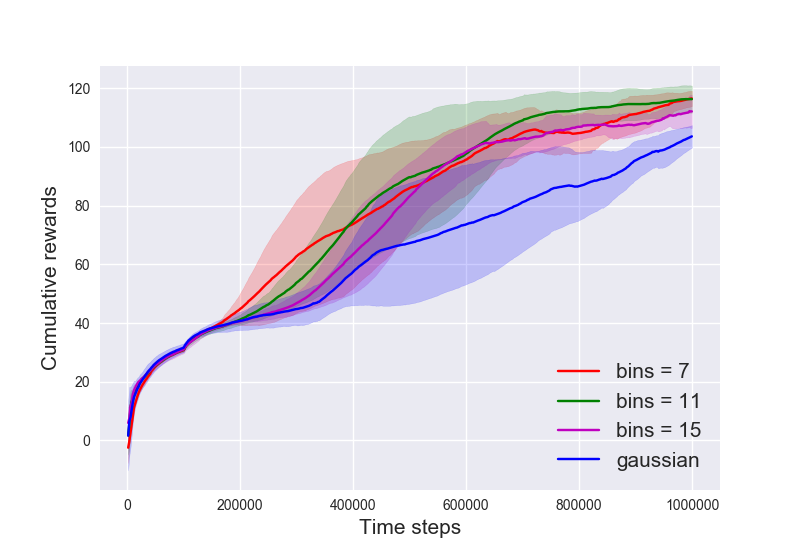}}
\subfigure[\textbf{Inverted Pendulum}]{\includegraphics[width=.23\linewidth]{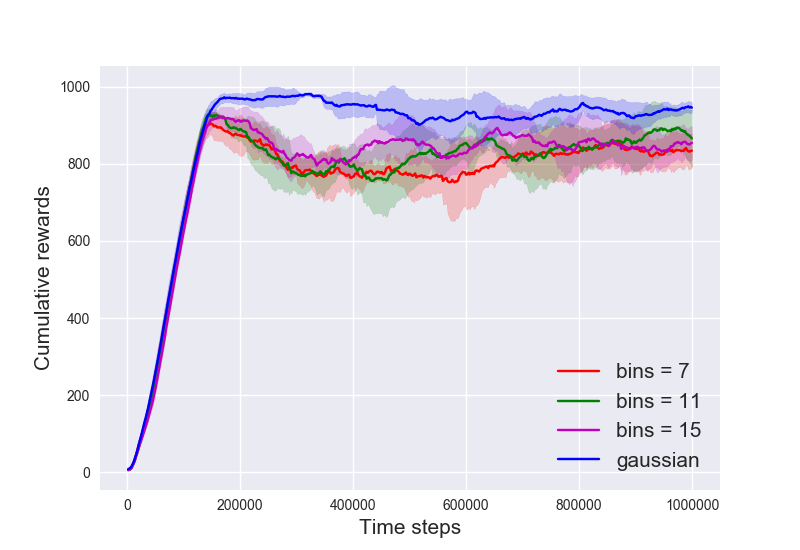}}
\subfigure[\textbf{Double Pendulum}]{\includegraphics[width=.23\linewidth]{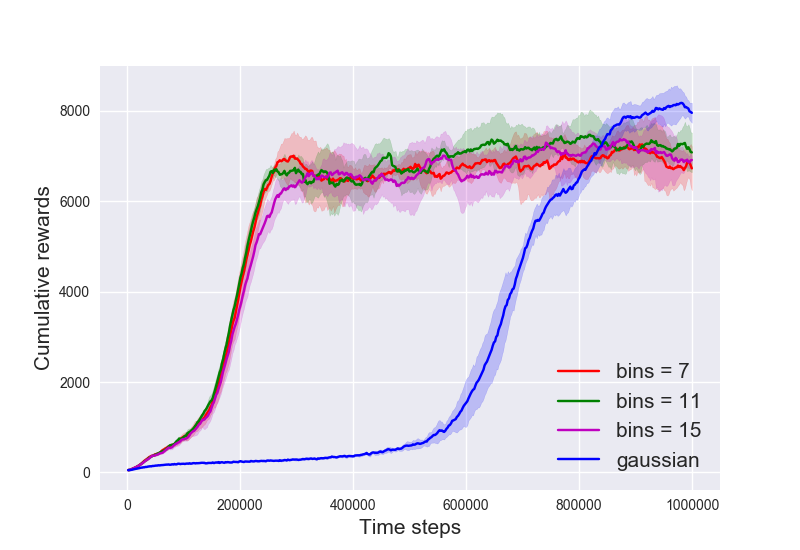}}
\caption{\small{MuJoCo Benchmarks: learning curves of PPO on OpenAI gym MuJoCo locomotion tasks. Each curve corresponds to a different policy architecture (Gaussian or discrete policy with varying number of bins $K=7,11,15$). Discrete policy significantly outperforms Gaussian on Humanoid tasks.}}
\label{figure:benchmarkppoappendix}
\end{figure}

We show results for PPO on simpler MuJoCo tasks in Figure \ref{figure:benchmarkppoappendix}. In such tasks, discrete policy does not necessarily outperform factorized Gaussian.

\begin{figure}[h]
\centering
\subfigure[\textbf{Bipedal Walker}]{\includegraphics[width=.45\linewidth]{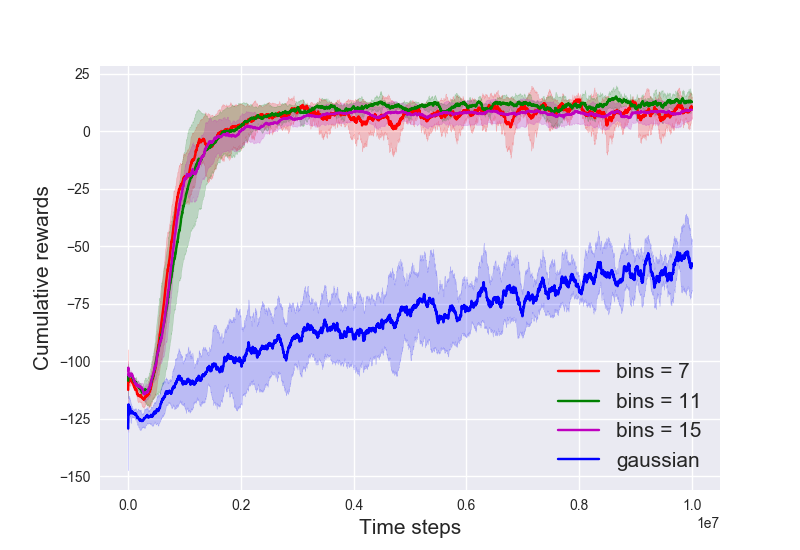}}
\subfigure[\textbf{Lunar Lander}]{\includegraphics[width=.45\linewidth]{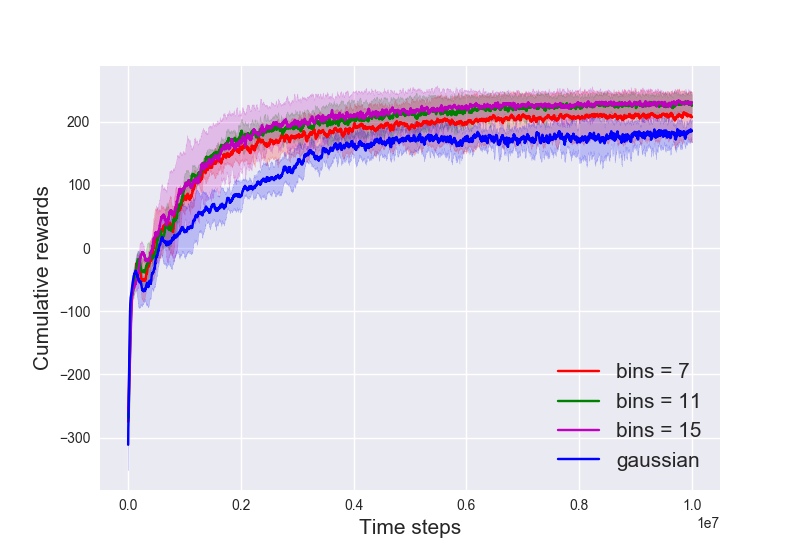}}
\caption{\small{Box2D Benchmarks: learning curves of PPO on OpenAI gym Box2D locomotion tasks. Each curve corresponds to a different policy architecture (Gaussian or discrete policy with varying number of bins $K=7,11$).}}
\label{figure:benchmarkppobox2d}
\end{figure}

\subsection{TRPO}
\begin{figure}[h]
\centering
\subfigure[\textbf{Reacher}]{\includegraphics[width=.23\linewidth]{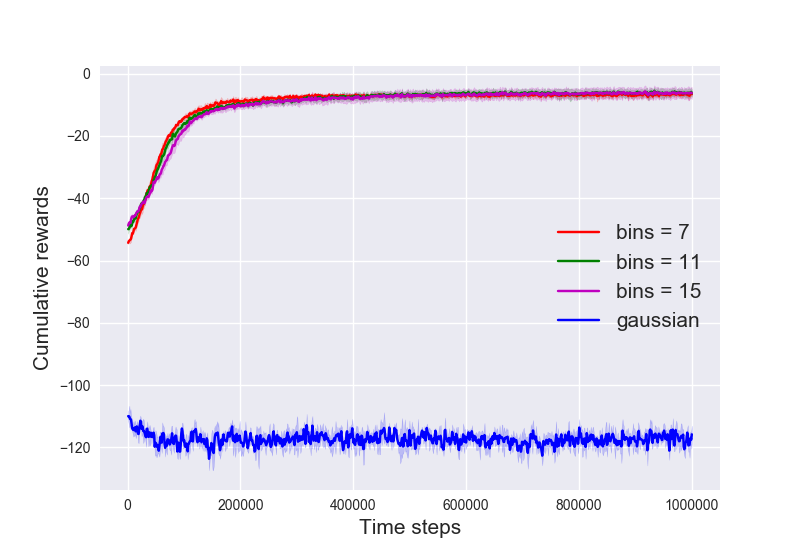}}
\subfigure[\textbf{Swimmer}]{\includegraphics[width=.23\linewidth]{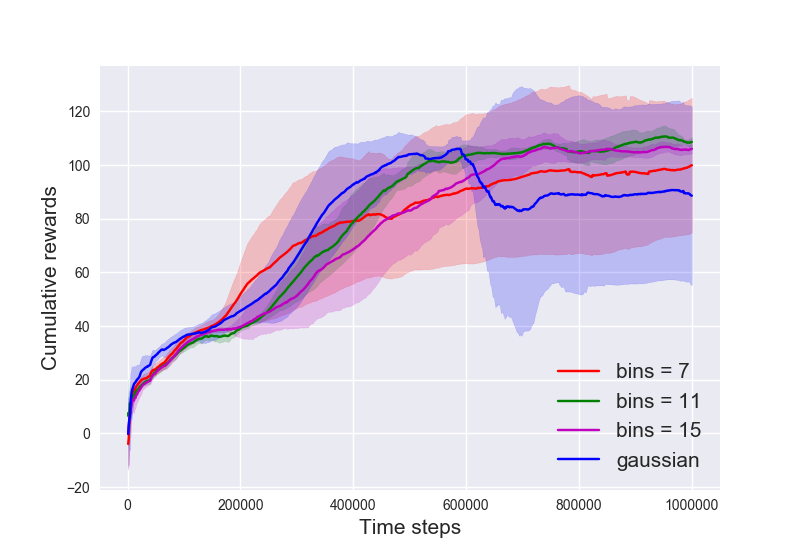}}
\subfigure[\textbf{Inverted Pendulum}]{\includegraphics[width=.23\linewidth]{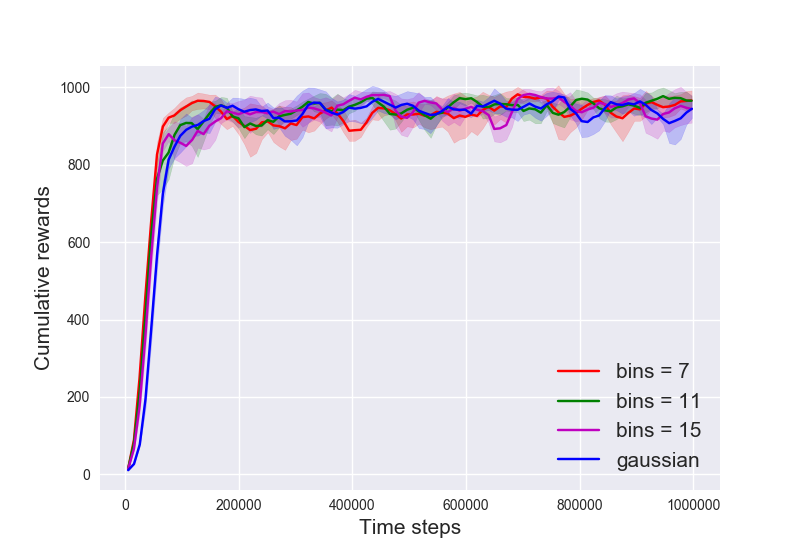}}
\subfigure[\textbf{Double Pendulum}]{\includegraphics[width=.23\linewidth]{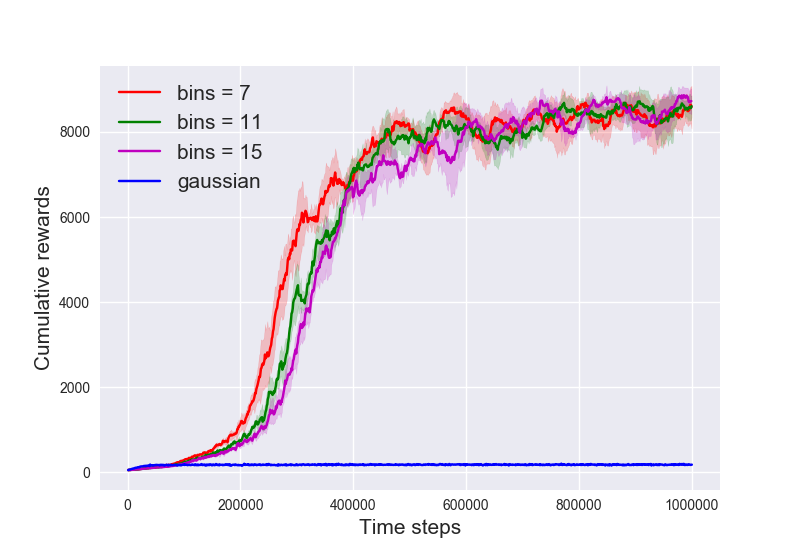}}
\caption{\small{MuJoCo Benchmark : learning curves of TRPO on MuJoCo locomotion tasks. Each curve is averaged over 6 random seeds and shows $\text{mean} \pm \text{std}$ performance. Each curve corresponds to a different policy representation (Red: Implicit, Green: GMM $K=5$, Yellow: GMM $K=2$, Blue: Gaussian). Vertical axis is the cumulative rewards and horizontal axis is the number of time steps.}}
\label{figure:benchmarktrpoappendix}
\end{figure}

We show results for TRPO on simpler MuJoCo tasks in Figure \ref{figure:benchmarktrpoappendix}. Even with simple task, discrete policy can still significantly outperform Gaussian ((a) Reacher and (d) Double Pendulum).

\begin{figure}[h]
\centering
\subfigure[\textbf{Bipedal Walker}]{\includegraphics[width=.45\linewidth]{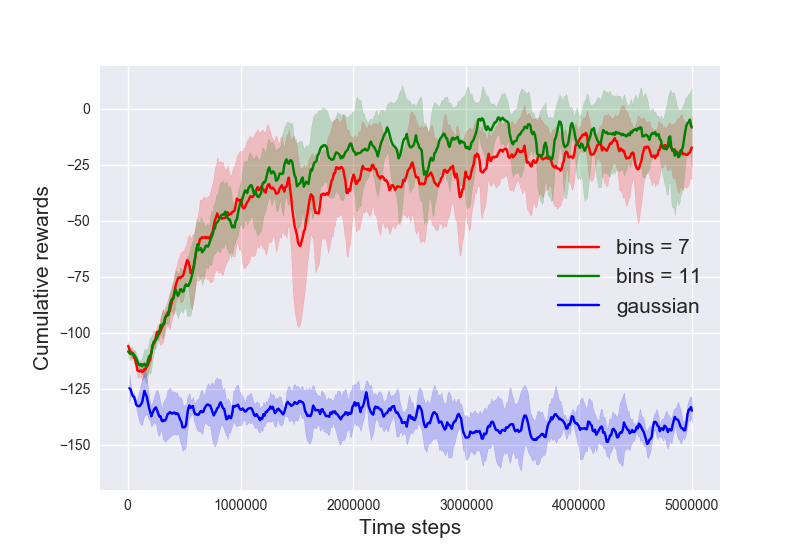}}
\subfigure[\textbf{Lunar Lander}]{\includegraphics[width=.45\linewidth]{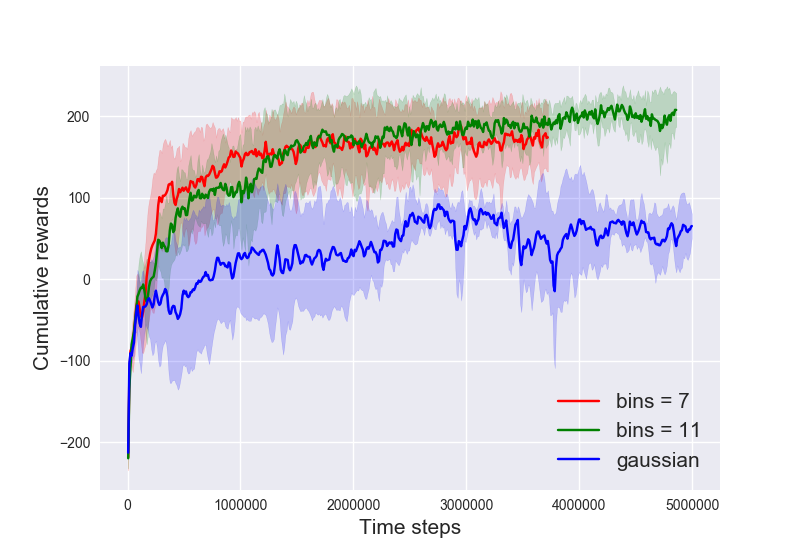}}
\caption{\small{Box2D Benchmarks: learning curves of TRPO on OpenAI gym Box2D locomotion tasks. Each curve is averaged over 5 random seeds and shows $\text{mean} \pm \text{std}$ performance. Each curve corresponds to a different policy architecture (Gaussian or discrete policy with varying number of bins $K=7,11$). Vertical axis is the cumulative rewards and horizontal axis is the number of time steps.}}
\label{figure:benchmarktrpobox2d}
\end{figure}

\subsection{ACKTR}
We show results for ACKTR on a set of MuJoCo and rllab tasks in Figure \ref{figure:benchmarkppoappendix}. For tasks with relatively simple dynamics, the performance gain of discrete policy is not significant ((a)(b)(c)). However, in Humanoid tasks, discrete policy does achieve significant performance gain over Gaussian ((d)(e)).

\begin{figure}[h]
\centering
\subfigure[\textbf{Reacher}]{\includegraphics[width=.3\linewidth]{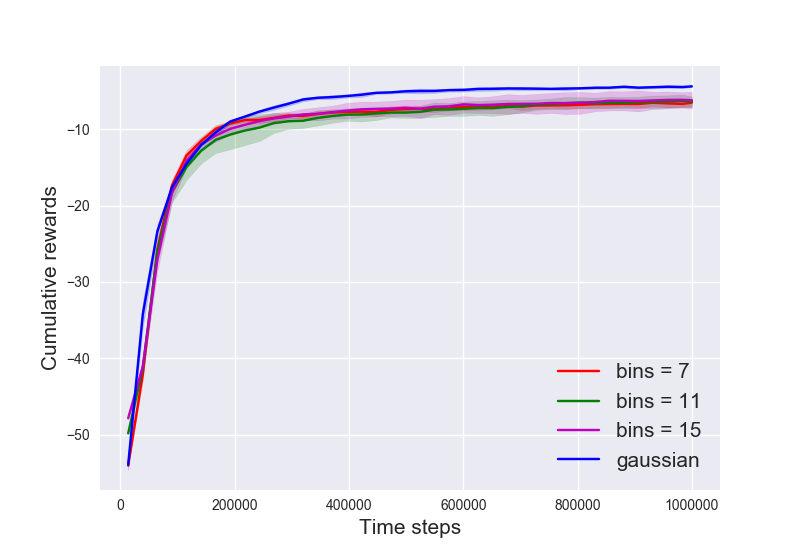}}
\subfigure[\textbf{Swimmer}]{\includegraphics[width=.3\linewidth]{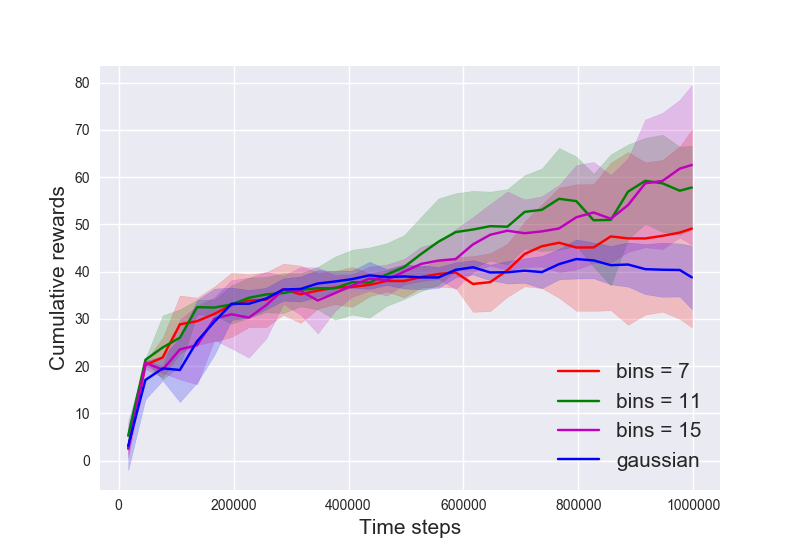}}
%\subfigure[Inverted Pendulum]{\includegraphics[width=.23\linewidth]{graph/benchmark_ppo_invertedpendulumpomdp}}
%\subfigure[Double Pendulum]{\includegraphics[width=.23\linewidth]{graph/%benchmark_ppo_inverteddoublependulumpomdp}}
%\subfigure[Hopper]{\includegraphics[width=.23\linewidth]{graph/benchmark_ppo_hopper}}
\subfigure[\textbf{HalfCheetah}]{\includegraphics[width=.3\linewidth]{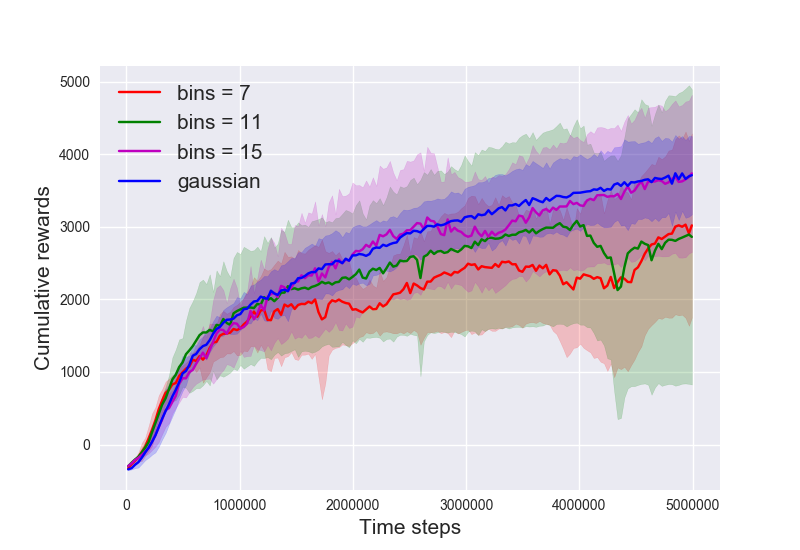}}
%\subfigure[Ant]{\includegraphics[width=.23\linewidth]{graph/benchmark_ppo_ant}}
%\subfigure[Walker]{\includegraphics[width=.23\linewidth]{graph/benchmark_acktr_walker2d}}
\subfigure[\textbf{Humanoid} (R)]{\includegraphics[width=.3\linewidth]{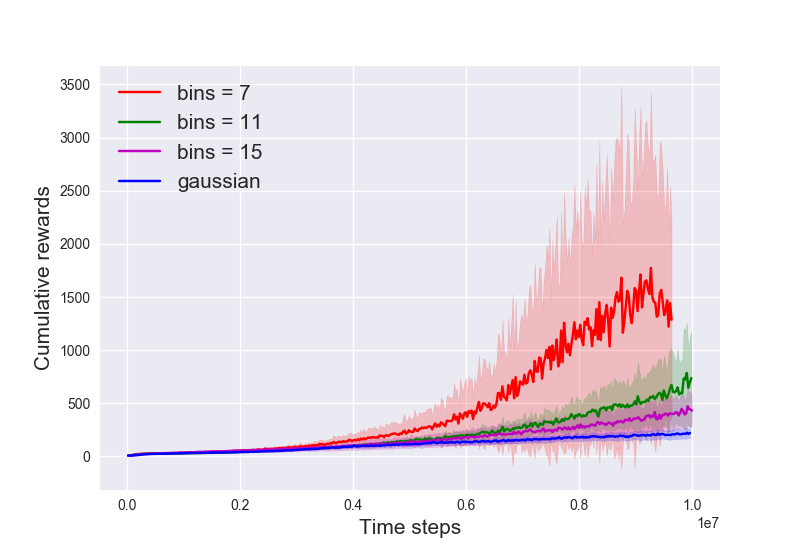}}
\subfigure[\textbf{Sim. Humanoid} (R)]{\includegraphics[width=.3\linewidth]{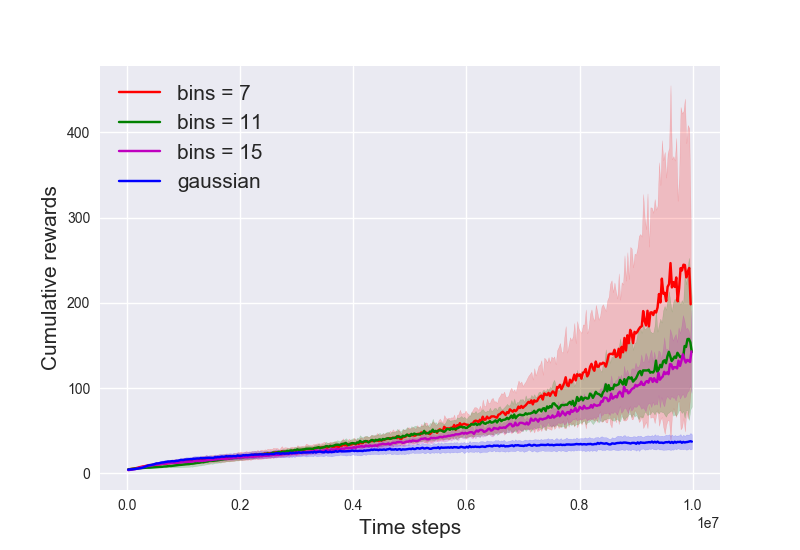}}
\subfigure[\textbf{Humanoid}]{\includegraphics[width=.3\linewidth]{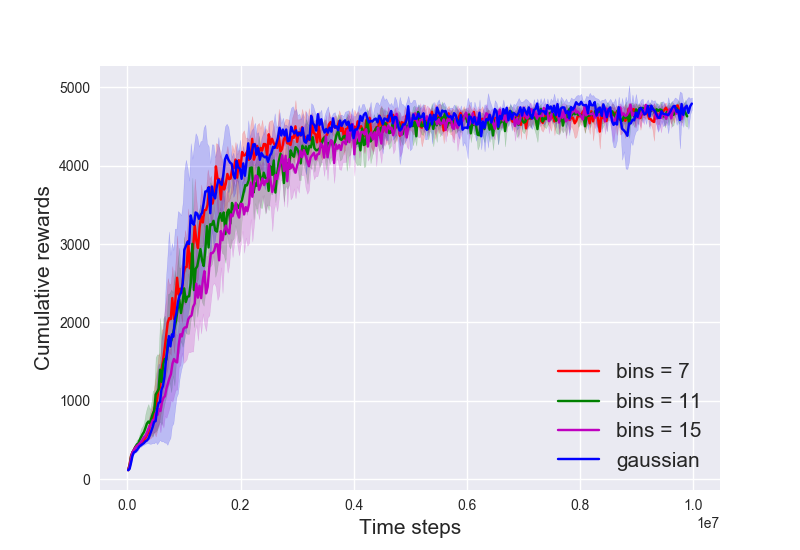}}
%\subfigure[HumanoidStandup]{\includegraphics[width=.23\linewidth]{graph/benchmark_ppo_humanoidstandup}}
\caption{\small{MuJoCo Benchmarks: learning curves of ACKTR on OpenAI gym MuJoCo locomotion tasks. Each curve is averaged over 5 random seeds and shows $\text{mean} \pm \text{std}$ performance. Each curve corresponds to a different policy architecture (Gaussian or discrete policy with varying number of bins $K=7,11,15$). Vertical axis is the cumulative rewards and horizontal axis is the number of time steps. Discrete policy significantly outperforms Gaussian on Humanoid tasks. Tasks with (R) are from rllab.}}
\label{figure:benchmarkdist}
\end{figure}

\subsection{Sensitivity to Hyper-parameters}
We present the sensitivity results in Figure \ref{figure:pposensitivity} below.

\begin{figure}[h]
\centering
\subfigure[\textbf{Roboschool Reacher}]{\includegraphics[width=.23\linewidth]{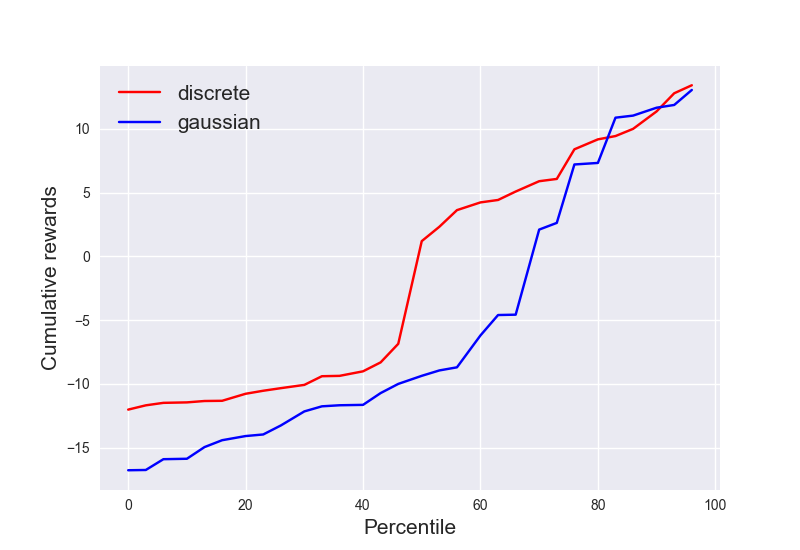}}
\subfigure[\textbf{Hopper}]{\includegraphics[width=.23\linewidth]{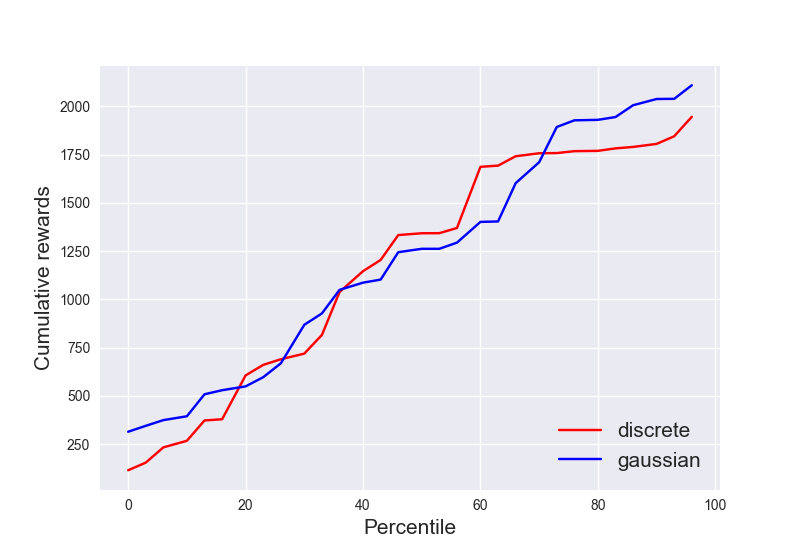}}
\subfigure[\textbf{HalfCheetah}]{\includegraphics[width=.23\linewidth]{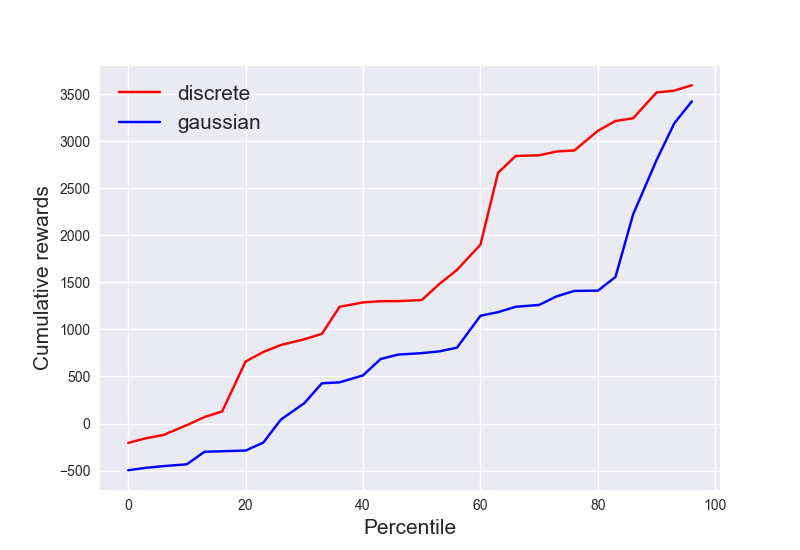}}
\subfigure[\textbf{Roboschool Ant}]{\includegraphics[width=.23\linewidth]{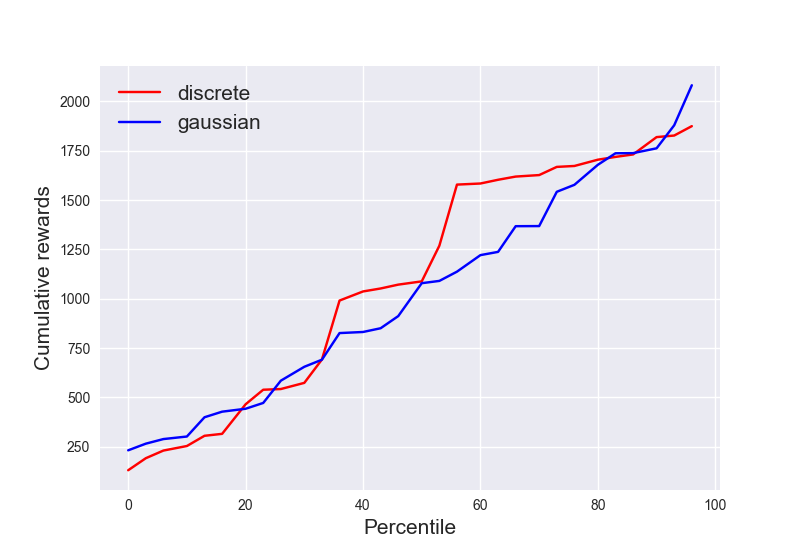}}
\subfigure[\textbf{Roboschool Walker}]{\includegraphics[width=.23\linewidth]{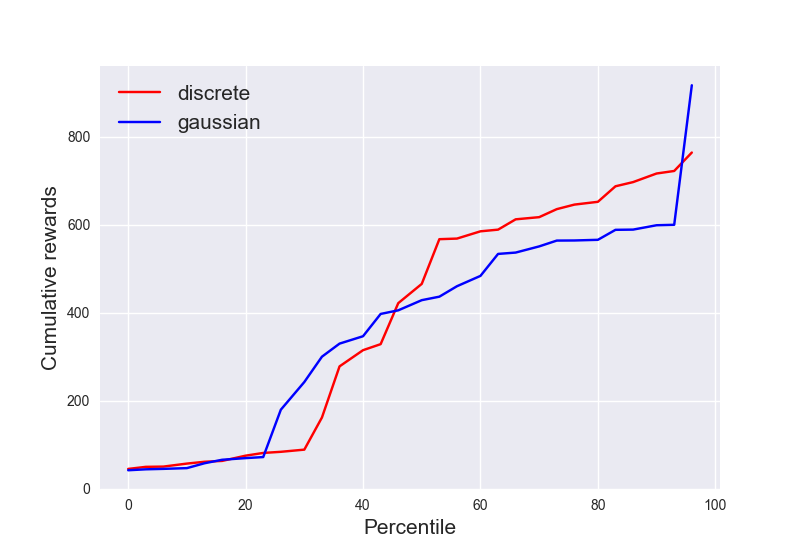}}
\subfigure[\textbf{Humanoid}]{\includegraphics[width=.23\linewidth]{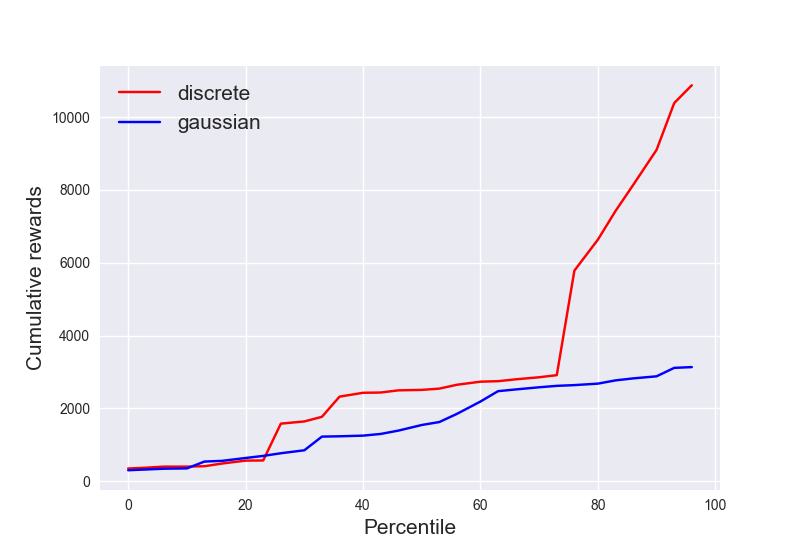}}
\subfigure[\textbf{Sim. Humanoid} (R)]{\includegraphics[width=.23\linewidth]{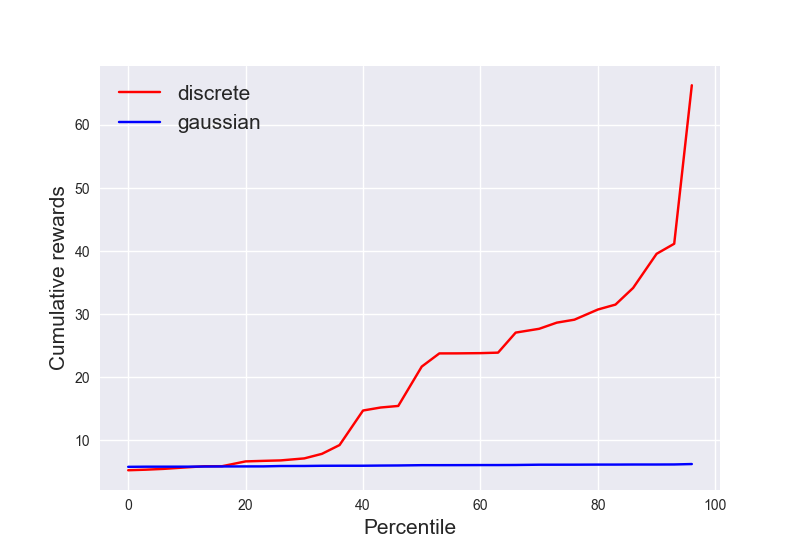}}
\subfigure[\textbf{Humanoid} (R)]{\includegraphics[width=.23\linewidth]{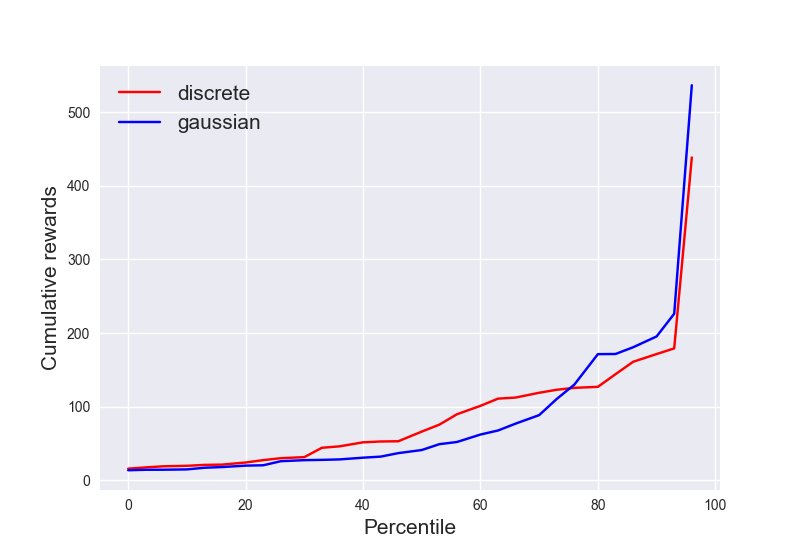}}
%\subfigure[HumanoidStandup]{\includegraphics[width=.23\linewidth]{graph/benchmark_ppo_humanoidstandup}}
\caption{\small{PPO sensitivity: quantile plots of final performance on benchmark tasks in OpenAI MuJoCo, rllab and Roboschool. In each plot, 30 different hyper-parameters are drawn for each policy (Gaussian vs. discrete policy). Reacher is trained for $10^6$ steps, Hopper $2\cdot 10^6$ steps and all other tasks about $5 \cdot 10^6$ steps. Tasks with (R) are from rllab.}}
\label{figure:pposensitivity}
\end{figure}

\subsection{Comparison with Gaussian Policy with Big Networks}
In our implementation, discrete/ordinal policy have more parameters than Gaussian policy. A natural question is whether the gains in policy optimization is due to a bigger network. To test this, we train Gaussian policy with large networks: 2-layer neural network with $128$ hidden units per layer. In Table \ref{table:trponetworksize} and Table \ref{table:pponetworksize}, we find that for Gaussian policy, bigger network does not perform as well as the smaller network  ($32$ hidden units per layer). Since Gaussian policy with bigger network has more parameters than discrete policy, this validates the claim that the performance gains of discrete policy policy are not (only) due to increased parameters. Below in Table \ref{table:pponetworksize} we show results for PPO and Table \ref{table:trponetworksize} for TRPO.

\begin{table}[t]
\caption{Comparison of TRPO $+$ Gaussian policy with networks of different sizes. Big network has $128$ hidden units per layer while small network has $32$ hidden units per layer. Both networks have $2$ layers. Small networks generally performs better than big networks. Below we show average $\pm$ std cumulative rewards after training for $5 \cdot 10^6$ steps.}
\label{policy}
\vskip 0.15in
\begin{center}
\begin{small}
\begin{sc}
\begin{tabular}{lcc}
\toprule
Task &  Gaussian (big) & Gaussian (small)\\
\midrule
Ant & $-104 \pm 30$ & $\mathbf{-94 \pm 44}$ \\
Sim. Human. (L) & $5.1 \pm 0.7$ & $\mathbf{6.4 \pm 0.4}$ \\
Humanoid & $501 \pm 14$ & $\mathbf{708  \pm 43}$ \\
Humanoid (L) & $20 \pm 2$ & $\mathbf{53 \pm 9}$ \\
\bottomrule
\end{tabular}
\end{sc}
\end{small}
\end{center}
\vskip -0.1in
\label{table:trponetworksize}
\end{table}

\begin{table}[t]
\caption{Comparison of PPO $+$ Gaussian policy with networks of different sizes. Big network has $128$ hidden units per layer while small network has $64$ hidden units per layer. Both networks have $2$ layers. Small networks generally performs better than big networks. Below we show averageg $\pm$ std cumulative rewards after training for $5 \cdot 10^6$ steps.}
\label{policy}
\vskip 0.15in
\begin{center}
\begin{small}
\begin{sc}
\begin{tabular}{lcc}
\toprule
Task &  Gaussian (big) & Gaussian (small)\\
\midrule
Ant & $\mathbf{3712 \pm 315}$ & $3317 \pm 152$ \\
Sim. Human. (L) & $4.8 \pm 0.3$ & $\mathbf{4.9 \pm 0.4}$ \\
Humanoid & $3221 \pm 535$ & $\mathbf{3766 \pm 413}$ \\
Humanoid (L) & $499 \pm 285$ & $\mathbf{1626 \pm 1480}$ \\
\bottomrule
\end{tabular}
\end{sc}
\end{small}
\end{center}
\vskip -0.1in
\label{table:pponetworksize}
\end{table}

\subsection{Additional Comparison with Beta policy}
\citet{chou2017improving} show the performance gains of Beta distribution policy for a limited number of benchmark tasks, most of which are relatively simple (with low dimensional observation space and action space). However, they show performance gains on Humanoid-v1. We compare the results of our Figure \ref{figure:benchmarkdist} with Figure 5(j) in \citep{chou2017improving} (assuming each training epoch takes $\approx 2000$ steps): within 10M training steps, discrete/ordinal policy achieves 
faster progress, reaching $\approx 4000$ at the end of training while Beta policy achieves $\approx 1000$. According to \citep{chou2017improving}, Beta distribution can have an asymptotically better performance with $\approx 6000$, while we find that discrete/ordinal policy achieves asymptotically $\approx 5000$.

\section{Illustration of Benchmark Tasks}
\begin{figure*}[h]
\centering
\includegraphics[width=.9\linewidth]{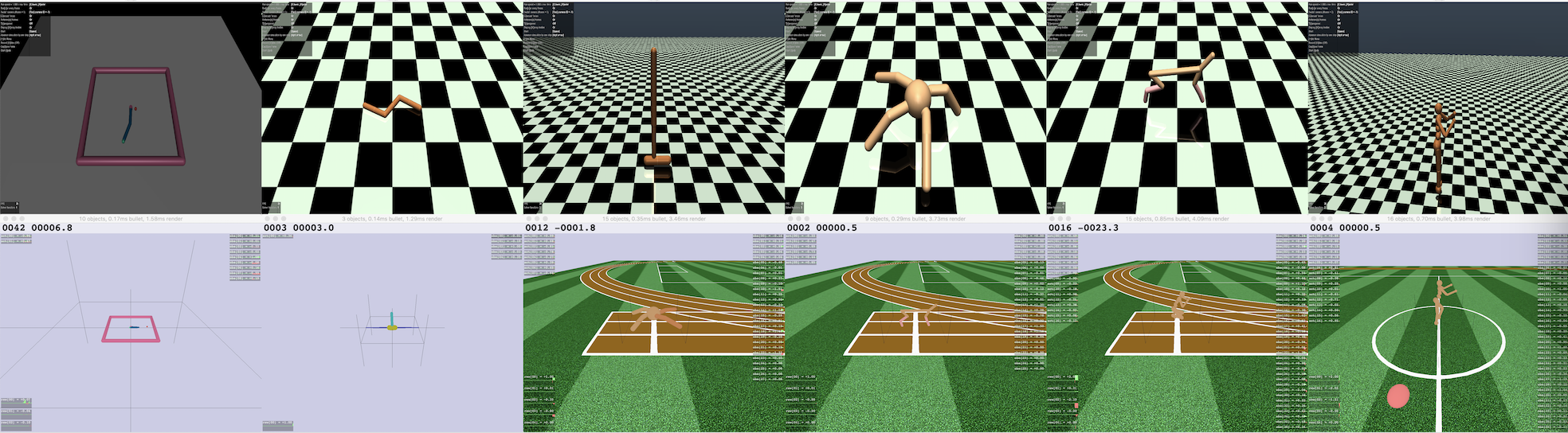}
\caption{\small{Benchmark Tasks: Illustration of locomotion benchmark tasks in OpenAI gym \citep{brockman2016}, rllab \citep{duanxi2016} with MuJoCo \citep{todorov2008} as simulation engines (top row) and Roboschool with open source simulation engine \citep{schulman2017} (bottom row). All tasks involve using sensory data as states and actuator controls as actions.}}
\label{figure:mujoco}
\end{figure*}

We present an illustration of benchmark tasks in Figure \ref{figure:mujoco}. All the benchmark tasks are implemented with very efficient physics simulators. All tasks use sensory data as states and actuator controls as actions.

\end{document}